\documentclass[preprint,12pt,authoryear,times]{elsarticle}

\usepackage{amssymb}
\usepackage{amsmath}
\usepackage{array}
\usepackage{chngcntr}
\usepackage{caption} 
\usepackage{calc}
\usepackage{lineno}
\usepackage{threeparttable}
\usepackage[colorlinks=true, linkcolor=blue, citecolor=blue, urlcolor=blue]{hyperref}
\usepackage{booktabs}
\usepackage{enumitem}
\usepackage{tabularx}
\usepackage{multirow}
\usepackage{natbib}
\usepackage{array}
\usepackage{float}

\journal{Elsevier}

\usepackage{url}
\usepackage{boxedminipage}
\usepackage{textpos}

\begin{document}

\begin{frontmatter}

\title{City landscape in sight: A crowdsourced framework for unlocking urban-scale window view perceptions from real estate imagery}

\affiliation[inst1]{organization={Department of Architecture, National University of Singapore},
            country={Singapore}}

\affiliation[inst2]{organization={College of Horticulture and Forestry Sciences / Hubei Engineering Technology Research Centre for Forestry Information, Huazhong Agricultural University},
            city={Wuhan},
            postcode={430070}, 
            country={China}}

\affiliation[inst3]{organization={School of Engineering and Applied Science, University of Pennsylvania},
            city={Philadelphia},
            postcode={19104},
            country={USA}}

\affiliation[inst4]{organization={Department of Political Science, Rutgers University},
            city={Newark},
            postcode={07102},
            country={USA}}

\affiliation[inst5]{organization={
School of Arts and Communication, China University of Geosciences},
            city={Wuhan},
            postcode={430070}, 
            country={China}}
            
\affiliation[inst6]{organization={Department of Real Estate, National University of Singapore},
            country={Singapore}}

\author[inst1,inst2]{Chucai Peng\fnref{equal}}
\author[inst1,inst3]{Sijie Yang\fnref{equal}}
\author[inst4]{Ang Liu}
\author[inst5]{Yang Xiang}
\author[inst2]{Zhixiang Zhou}
\author[inst1,inst6]{Filip Biljecki\corref{cor1}}
\ead{filip@nus.edu.sg}

\fntext[equal]{These authors contributed equally to this work.}
\cortext[cor1]{Corresponding author.}

\begin{abstract}
\begin{textblock*}{\textwidth}(0cm,-14.4cm)
\begin{center}
\begin{footnotesize}
\begin{boxedminipage}{1\textwidth}
This is the Accepted Manuscript version of an article published in the journal \emph{Landscape and Urban Planning} in 2026, which is available at: \url{https://doi.org/10.1016/j.landurbplan.2026.105734}\\ Cite as:
Peng C, Yang S, Liu A, Xiang Y, Zhou Z, Biljecki F (2026): City landscape in sight: A crowdsourced framework for unlocking urban-scale window view perceptions from real estate imagery. \textit{Landscape and Urban Planning}, 275: 105734.
\end{boxedminipage}
\end{footnotesize}
\end{center}
\end{textblock*}
City landscapes viewed through home windows influence quality of life, yet perceptions of actual window views at the urban scale remain understudied. This study presents an approach for large-scale mapping of perceptions using 12,334 window view images (WVIs) collected from actual residential properties listed on real estate platforms in Wuhan, China, representing a rarely explored form of urban view imagery that offers advantages over the rendered or simulated window views commonly examined in previous studies. Through a non-immersive virtual reality platform, we collected 27,477 pairwise comparisons across six perceptual dimensions (e.g.\ preference) from 304 participants based on 499 WVIs. A hybrid neural network model was trained to predict human perceptions of all crowdsourced WVIs and map their spatial distribution. Results reveal significant spatial autocorrelation with distinct hot and cold spots across the whole city. Floor level strongly influences human perceptions: while higher floors offer more preferred and extensive window views, lower-floor windows provide residents with quiet and vivid views. An inference model further shows that window view composition matters considerably: high ratios of sky, trees, and low-rise buildings enhance people's preferences and perceptions of vividness, whereas high ratios of high-rise buildings increase perceptions of monotony and oppression. Importantly, these effects are non-linear: the excessive presence of certain elements can alter their impact on human perception. This work advances urban-scale understanding of residents' visual experiences and offers a transferable, human-centric method to inform urban planning and design aimed at improving the visual quality of window views.
\end{abstract}




\end{frontmatter}


\section*{Nomenclature}
\begin{tabular}{@{}p{0.08\textwidth}p{0.36\textwidth}p{0.08\textwidth}p{0.38\textwidth}@{}}
WVI & Window view imagery & SHAP & Shapley Additive exPlanations \\
SVI & Street view imagery & VIF & Variance inflation factor \\
VR & Virtual reality & NDBI & Normalised Difference Built-up Index \\
NN & Neural network & NDVI & Normalised Difference Vegetation Index \\
RMSE & Root mean square error & VAUA & Visual assessment of urban affordance \\
\end{tabular}

\section{Introduction}


Windows serve as vital interfaces connecting residents to city landscapes beyond their homes \citep{an2019optimal, du2022impact, abdelrahman2023visible}. Through windows, residents experience diverse views of urban environments --- from green spaces and waterfronts to dense high-rise building clusters --- which profoundly shape their living experience and quality of life. In high-density urban environments, where direct access to nature is often limited, good views of the city landscape have been recognised for benefiting residents' psychological and physiological health \citep{olszewska-guizzo2018window, elsadek2020window}. High-quality views are linked to better life satisfaction in residential buildings, less stress, higher work productivity in offices, and even faster recovery in healthcare settings \citep{ulrich1984view, li2016impact, chang2020life, ko2021, lindemann-matthies2021associations}. Moreover, better window views are often associated with higher housing prices \citep{peng2025measuring}, highlighting their relevance across both life health and real estate economic domains. As a crucial part of the urban comfort experience, understanding how residents subjectively perceive the city landscapes through window views from their own houses is therefore important for future human-centred urban design and planning, which can be assessed through in-field comfort investigations and computational comfort modelling \citep{yang2025urbana}. It is worth noting that window view quality is a multi-dimensional concept, encompassing not only \textit{view content} (what occupants see through the window), but also \textit{view access} (how much of the window view is visible) and \textit{view clarity} (how clearly the view content can be perceived) \citep{ko2021}. This study focuses specifically on view content, as it is the dimension most amenable to large-scale assessment through crowdsourced real estate imagery.

Given these multifaceted benefits, ensuring access to high-quality window views through urban design and planning is a promising approach to improving citizens' health and well-being in urban environments. However, despite the growing recognition of the importance of window views, existing studies remain limited in scope and depth. Prior studies 1)often focus on singular perceptual dimensions such as preference or oppressiveness, 2)are constrained to small-scale datasets or simulation-based settings, rather than relying on real scenarios, and 3)lack citywide coverage necessary to inform urban design and planning practice \citep{li2022room}. Thus, there are clear gaps in understanding how multidimensional human perceptions of window views are distributed across a city and which city landscape factors drive these perceptions.

These limitations are primarily due to the challenges of acquiring both real-world window view imagery (WVI) and human perceptual data at a large scale \citep{li2024cim}. Conventional methods, such as field surveys or computer simulations, are labour-intensive and struggle to capture the detailed visual and spatial texture of real-world urban environments separately \citep{schmid2021outlook, lin2022evaluation}. Meanwhile, some real estate listing platforms have emerged as promising sources of large-scale real estate data, including property imagery (e.g., photos of kitchens in flats, building amenities, and floor plans). A subset of such imagery often includes WVIs, which can reflect residents' actual window views across diverse city locations, even in areas with different building heights. Such datasets offer the potential for large-scale window view studies based on real images, which can be leveraged to investigate and map residents' subjective experiences of city landscapes through their windows.

In this study, we use crowdsourced real estate images and address these gaps by introducing a comprehensive framework for investigating, modelling, and mapping subjective perceptions of window views at the urban scale. Using 12,334 actual window view images collected from property listing platforms in Wuhan, China, we modelled human perception based on online pairwise survey data obtained through a non-immersive virtual reality (VR) interface (27,477 valid pairwise comparisons from 304 participants based on 499 selected WVIs). Our framework enables the prediction and inference modelling of six human perceptual dimensions --- preference (\textit{Prefer}), monotony (\textit{Monotonous}), quietness (\textit{Quiet}), extensiveness (\textit{Extensive}), vividness (\textit{Vivid}), and oppressiveness (\textit{Oppressive}) --- across citywide window views. It is a dual-crowdsourced framework, as depicted in \ref{fig_concept}, where real estate imagery is crowdsourced from myriads of real estate agents and property owners, while human perception data are crowdsourced from hundreds of participants in window view perception surveys. All this data allows us to model window view perception, make urban-scale predictions, and map its spatial distribution across the whole city to uncover potential hot and cold spots and relationships between visual composition (sky, trees, buildings) and human perception.

In summary, this study makes the following key contributions:

\begin{itemize}
    \item We establish a fully crowdsourced framework for urban window view perception research, leveraging both crowdsourced real estate imagery (uploaded by property agents and sellers/landlords) and crowdsourced human perception data (collected via online pairwise comparisons by survey participants). This dual-crowdsourced approach demonstrates the utility of real estate advertisements --- a rarely exploited urban sensing data source --- for subjective urban perception research, expanding their potential use cases beyond property valuation. Focusing predominantly on crowdsourced data offers a human-centric approach and increases the role of public-contributed data in this domain. Likewise, a secondary contribution of this work is in the geospatial domain, giving more attention to this rarely used form of user-generated geographic information (or Volunteered Geographic Information -- VGI).
    \item We develop a hybrid neural network (NN) model for perception prediction and an inference model to decode how built environment characteristics (semantic segmentation, land cover, urban form) shape multidimensional subjective experiences.
    \item We systematically map urban-scale spatial patterns of perceptions, revealing significant spatial autocorrelation, distinct hotspots and cold spots, and vertical perception gradients across floor levels.
    \item We identify non-linear relationships between visual composition and perceived quality: while sky and trees generally enhance preference and vividness, excessive presence can diminish quality; high-rise buildings consistently increase monotony and oppression.
    \item Results from this study can inform human-centred urban planning, including how built environment characteristics relate to the visual environments visible from residents' windows.
\end{itemize}

\section{Related work}
\subsection{Data collection and research methods in window view research}

\begin{figure}[ht]
    \centering
    \includegraphics[width=1\textwidth]{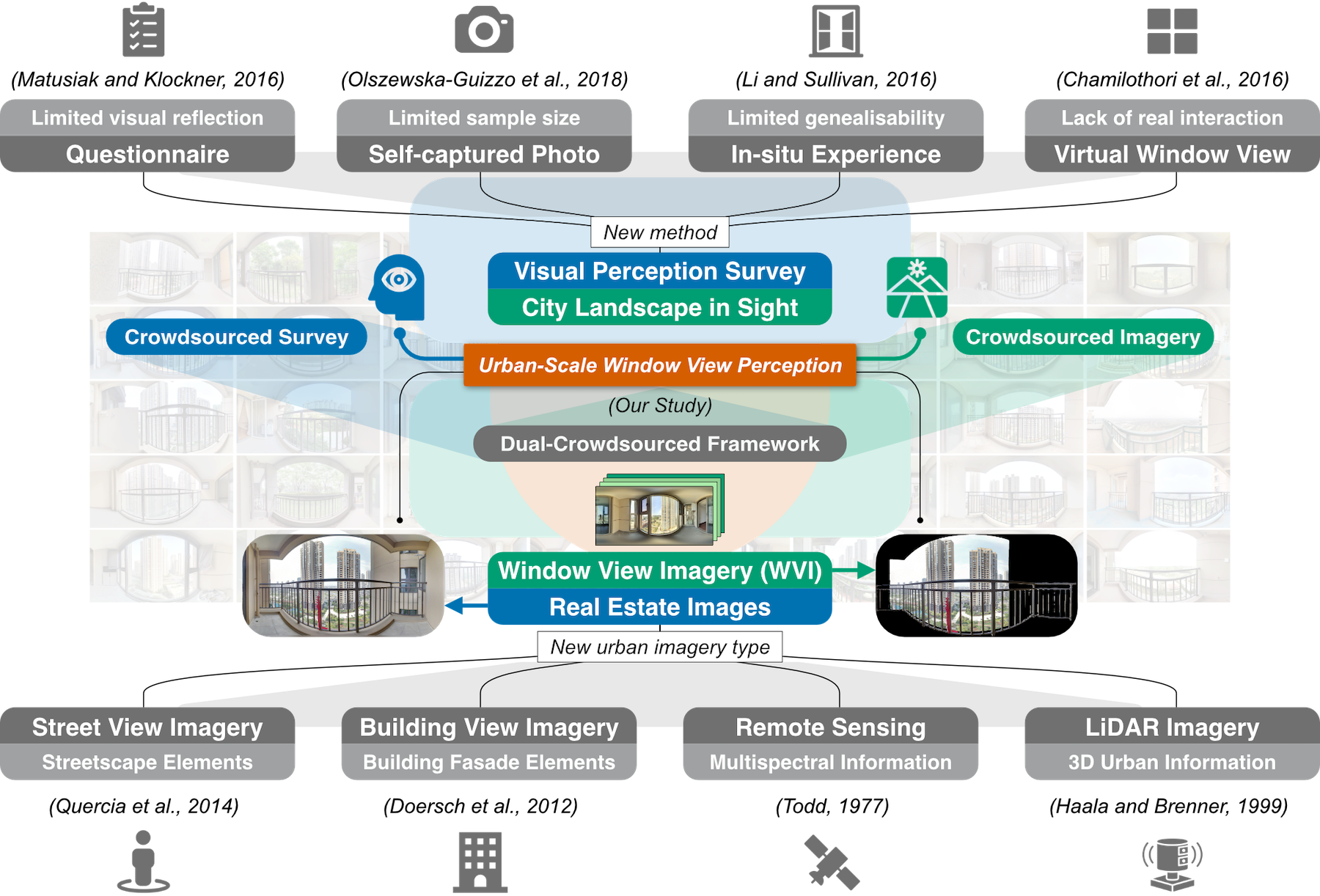}
    \caption{Conceptual framework of this study together with traditional approaches and example references.}
    \label{fig_concept}
\end{figure}

Understanding how residents perceive city landscapes through their windows requires appropriate data sources that balance authenticity, scale, and perceptual richness. Window view studies have adopted diverse data collection approaches, which can be broadly categorised into five types: questionnaire-based surveys, self-captured photographs, in-situ experience studies, and two types of simulations --- virtual window views with controlled variables and city-scale simulations (Table \ref{tab:window_view_studies}, Figure \ref{fig_concept} upper panel).

\begin{table}[ht]
    \centering
    \scriptsize
    \renewcommand{\arraystretch}{1.15} 
    \caption{Comparison of window view data sources.}
    \label{tab:window_view_studies}
    \begin{tabularx}{\textwidth}{p{2.4cm}*{7}{>{\centering\arraybackslash}X}}
    \toprule
    Data Source & Authenticity & Scalability & City-scale & Diversity & Visual-rich & Controlled & Cost-effective \\
    \midrule
    Questionnaire & & $\bullet$ & & $\bullet$ & & & $\bullet$ \\
    Self-captured photos & $\bullet$ & $\bullet$ & & & $\bullet$ & & $\bullet$ \\
    In-situ experience & $\bullet$ & & & & $\bullet$ & $\bullet$ & \\
    Virtual (controlled) & & & & & & $\bullet$ & $\bullet$ \\
    Virtual (simulation) & & $\bullet$ & $\bullet$ & & & & $\bullet$ \\\midrule
    \textbf{Real estate imagery} & $\bullet$ & $\bullet$ & $\bullet$ & $\bullet$ & $\bullet$ & $\bullet$ & $\bullet$ \\
    \bottomrule
    \multicolumn{8}{p{\textwidth}}{\scriptsize Note: $\bullet$ indicates the data source has the corresponding feature. Authenticity: window view data are captured in real-world scenes; Scalability: perception data can be collected at large-scale; City-scale: window view data can cover entire urban areas; Diversity: window view data provide diverse visual scenes and contexts; Visual-rich: window view data provide detailed and textured visual information; Controlled: data source can provide systematic control over environmental variables; Cost-effective: data collection is economic.} \\
    \end{tabularx}
\end{table}

\textbf{Questionnaire-based studies} are widely used because of their flexibility for perception surveys~\citep{matusiak2016how, batool2021window}, typically applied to examine correlations between window views and living outcomes such as residents' mental well-being or life satisfaction~\citep{korpela2017nature, chang2020life}. Recent works have improved perception questionnaires by incorporating user-submitted photos and computer vision analytics~\citep{hasegawa2022potential, zhang2023is}. While questionnaire-based methods offer scalability and cost-effectiveness, they often lack detailed and rich window view data for specific urban areas.

\textbf{Self-captured photographs} provide a more authentic record of residents' perspectives on window views than questionnaire stimuli or simulations, although the actual in-situ window view itself remains the most representative reference~\citep{olszewska-guizzo2018window, kent2020evaluation, schmid2021outlook, lin2022evaluation}. Like questionnaires, they are relatively low-cost and can be embedded in online questionnaires and shared broadly to reach large, geographically dispersed samples~\citep{zhang2023is, samaan2024low}, while additionally offering some control over external factors such as sunlight and glare through the choice of capture time. In practice, however, such studies still rely on participants photographing their own views, which constrains standardisation and tends to limit coverage to the participants' own residences rather than achieving systematic citywide sampling.

\textbf{In-situ experience studies} invite participants to directly experience real window views in controlled environments, often accompanied by physiological or psychological measurements using related devices~\citep{li2016impact, elsadek2020window, du2022impact, ko2022window, yao2024natural}. These studies are effective at exploring causal relationships between landscape features of window views and human perception survey results, as they capture nuanced human perceptual responses; however, their application is limited to very small scales, often limited to campus or laboratory settings with student populations.

\textbf{Virtual window views} have gained much attention with recent advances in computer vision and modelling. One stream emphasises experimental control using software such as Unity or Unreal Engine to create synthetic window views with highly adjustable parameters~\citep {chamilothori2019adequacy, moscoso2021window, chung2022study, wang2024exploring, ingabo2025contextual}, offering strong experimental control and cost-effectiveness. When such scenes are delivered through immersive virtual reality headsets, the reduction in authenticity is modest --- arising mainly from the imperfect rendering of conditions such as daylight and luminance contrast --- because headsets provide a more immersive experience than static images; this immersion, however, comes at the cost of greater resource requirements and unique problems such as simulation sickness~\citep{martirosov2022cyber, ugur2024brightness}. Another stream focuses on city-scale studies by simulating building window views using large-scale 3D city models~\citep{li2020new, li2022room, li2024cim}. While achieving broader spatial coverage and a highly flexible experimental design, these virtual simulations depend on model precision and may omit unique urban features, such as sunlight and other weather conditions, that occur in reality.

Despite methodological diversity, existing window view perception studies face persistent limitations, as we mentioned above (Table \ref{tab:window_view_studies}), such as small-scale or laboratory-based data, restricted spatial coverage and generalizability~\citep{schmid2021outlook, lin2022evaluation, ko2022window}, and a lack of authenticity in actual residential views~\citep{li2022room, kim2022seemo}. Moreover, most work focuses on limited perceptual dimensions --- typically preference or oppressiveness. Recent advances in apartment-level greenery measurement~\citep{das2025greenery} show promise; yet, gaps remain in understanding multidimensional perceptions at the urban scale. These constraints highlight a critical research gap: the need for data sources that simultaneously achieve authenticity, scalability, and urban-scale coverage. Recent work has further highlighted that assessment methods are often aligned with specific dimensions of view quality: VR is suited to evaluating view access, physical spaces to view clarity, and digital images to view content \citep{kim2025window}. Given that this study focuses on perceived view content at the urban scale, real estate imagery—comprising large volumes of in-situ window view images—provides a particularly appropriate data source. It enables citywide investigations while maintaining both visual richness and real-world authenticity, thereby addressing key limitations of existing approaches.

\subsection{Urban imagery typology and window view imagery as promising perspective}

\begin{table}[ht]
    \centering
    \scriptsize
    \renewcommand{\arraystretch}{1.15} 
    \caption{Complementary characteristics of urban data types.}
    \label{tab:urban_imagery}
    \begin{tabularx}{\textwidth}{p{2.2cm}*{6}{>{\centering\arraybackslash}X}}
    \toprule
    Data Type & Perspective & Primary Use & Perception & Public & Residential & Vertical \\
    \midrule
    Street view & Pedestrian & Public space & $\bullet$ & $\bullet$ & & \\
    Building view & Facade & Architecture & $\bullet$ & $\bullet$ & & \\
    Remote sensing & Aerial & Land use & & $\bullet$ & & \\
    LiDAR / 3D city models & 3D / Variable & Built form & $\bullet$ & $\bullet$ & & \\\midrule
    \textbf{Window view} & \textbf{Interior} & \textbf{Residential} & $\bullet$ & & $\bullet$ & $\bullet$ \\
    \bottomrule
    \multicolumn{7}{p{\textwidth}}{\scriptsize Note: $\bullet$ indicates the primary strength. Perception: suitable for perception studies; Public: public spaces; Residential: residential interiors; Vertical: building height differentiation. Each imagery type serves complementary purposes.} \\
    \end{tabularx}
\end{table}

Urban research has leveraged diverse imagery types to understand city environments over the past decades, each capturing distinct perspectives that together form a complementary ecosystem of urban sensing (Figure \ref{fig_concept} lower panel; Table \ref{tab:urban_imagery}). These perspectives complement each other and support urban analytics approaches, including geometry-based~\citep{xu2022comparinga} and graph-based modelling~\citep{yap2025revealinga}.

\textbf{Street view imagery (SVI)} is the most widely adopted urban imagery type, capturing pedestrian-level perspectives of public spaces~\citep {biljecki2021street, ito2024understanding, ito2025zensvi}. SVI has been extensively used to assess urban perception, including aesthetic quality, safety, and environmental characteristics~\citep{quercia2014aesthetic, kang2020review, hou2022comprehensive}. Large-scale datasets spanning global cities have enabled cross-cultural studies~\citep{hou2024global}, while recent work demonstrates SVI's capacity to extract built environment features and quantify subjective perceptions~\citep{kang2018building, zhang2018measuring, qiu2022subjective, yang2023role}. However, comparative studies have revealed inherent biases~\citep{huang2025no, fan2025coverage}. SVI excels at perception studies of public spaces but focuses on publicly accessible environments.

\textbf{Building view imagery} focuses on architectural facades, characterising urban architecture and building aesthetics. Studies have employed building view imagery and computational methods to extract architectural elements~\citep{doersch2015makes}, evaluate human perception of building exteriors~\citep{liang2024evaluating}, and enrich public building data in open maps through automated information prediction~\citep{liang2025openfacades}. This imagery type offers insights into buildings in cities but emphasises exterior appearances rather than interior living experiences.

\textbf{Remote sensing imagery}, including satellite imagery and aerial imagery, provides top-view perspectives for urban analytics~\citep{weng2012remote, liu2015monitoring}. Applications include land-use classification, land-change detection, local climate zone mapping, etc.~\citep{voogt2003thermal}. While excelling at macro-scale spatial and temporal pattern recognition, remote sensing imagery is not typically used for urban perception studies due to a lack of a human-scale perspective.

\textbf{LiDAR (Light Detection and Ranging)} is not considered a traditional imagery type but can offer precise three-dimensional morphological perspectives for the built environment and produce a series of image-like raster products (intensity, DEMs, DSMs)~\citep{haala1999extraction}. Applications include land cover classification~\citep{yan2015urban}, 3D building reconstruction~\citep{cheng20113d}, and urban visual quality assessment~\citep{wu2021mapping}. While LiDAR characterises public space geometry, it typically does not capture visual appearance, colour, or texture --- elements critical for perceptual studies.

\textbf{Window view imagery} is an emerging urban imagery type that captures city landscapes from real estate interior views --- the perspective from which residents take in real life. Window views have been proven related to human well-being, productivity, and stress reduction~\citep{ulrich1984view, elsadek2020window}, and contribute to property values~\citep{damigos2011value, peng2025measuring}. Unlike SVI, which captures pedestrian experiences in public street spaces, WVI is captured at viewpoints in private residential spaces across diverse floor levels. Investigating this perspective is particularly relevant and essential in high-density contexts where horizontal-vertical stratification creates dramatically different visual experiences for residents living at various places and floors. 


As highlighted in Figure \ref{fig_concept} and Table \ref{tab:urban_imagery}, WVI fills this gap through its distinctive viewpoints from building interiors. Rather than competing with existing urban imagery types, WVI complements them by addressing the visual quality of urban spaces within buildings. This study establishes WVI as a valuable addition to the urban imagery research ecosystem, demonstrating its potential for assessing citywide perceptions of window views.

\subsection{Computational modelling of subjective urban perception}

Understanding how people perceive urban environments has been a central topic in environmental psychology and urban studies for a long time. Early investigations focused on studying the effects of vegetation or natural elements in shaping people's psychological responses~\citep{smardon1988perception}, though these studies were limited to small sample sizes.

The emergence of crowdsourced data sources marked a shift, providing researchers with large-scale datasets for urban-scale human perception research. In 2013, \citeauthor{salesses2013collaborative} introduced the Place Pulse dataset, collecting millions of pairwise comparisons of SVIs to map human perceptions of streetscapes across global cities, demonstrating that urban perceptions could be systematically quantified at a large scale. The convergence of advanced prediction modelling methods and large-scale urban imagery datasets has since transformed the paradigm of urban perception research towards urban-scale prediction and studies~\citep{ito2024understanding}. In 2016, \citeauthor{dubey2016deep} trained deep learning models on crowdsourced SVI data to quantify six human perceptual dimensions across 56 cities worldwide. Subsequently, more and more studies began leveraging SVI and online SVI visual assessment surveys to predict multidimensional subjective urban experiences, revealing how visual elements in cities influence human perceptions~\citep{zhang2018measuring, qiu2022subjective, yang2023role, liang2024evaluating, yang2025thermal}.

However, existing computational perception research predominantly focuses on street-level public spaces with SVI, leaving residential visual environments from building interiors largely unexplored. There are some window view perception studies, but they have largely been limited to authenticity, survey scalability, or urban-scale coverage~\citep{schmid2021outlook, chung2022study, lin2022evaluation, wang2024assessing}, as mentioned in Section 2.1. The gap between well-developed streetscape perception research and current limitations in window view perception research offers a promising research direction based on WVI.

\section{Method}
\subsection{Research framework, study area and data collection}

\begin{figure}[ht]
    \centering
    \includegraphics[width=1\textwidth]{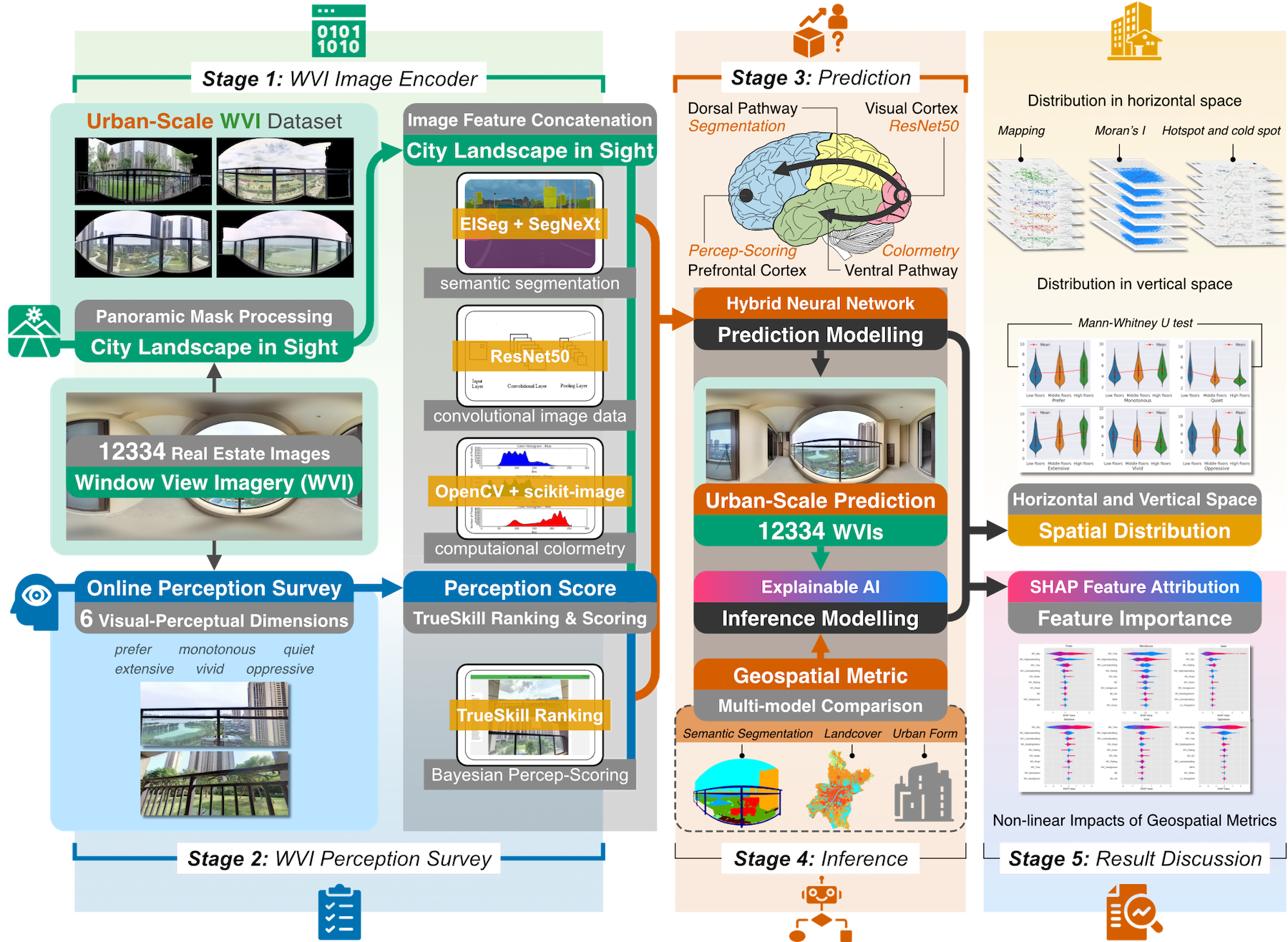}
    \caption{Five-stage research framework encompassing data collection, perception survey, prediction modelling, inference analysis, and result discussion.}
    \label{fig_framework}
\end{figure}

This study establishes a comprehensive five-stage analytical framework to map and understand subjective perceptions of window views at the urban scale (Figure \ref{fig_framework}). \textbf{Stage 1 (WVI Image Encoder)} involves panoramic mask processing and image feature concatenation, integrating semantic segmentation (using EISeg and SegNeXt), convolutional image data (ResNet50), and computational colourimetry (OpenCV and scikit-image). \textbf{Stage 2 (WVI Perception Survey)} implements an online perception survey across six visual-perceptual dimensions (Prefer, Monotonous, Quiet, Extensive, Vivid, Oppressive), employing TrueSkill ranking and Bayesian perception scoring to quantify human judgements. \textbf{Stage 3 (Prediction)} develops a hybrid neural network for prediction modelling, whose architecture integrates a dorsal pathway (spatial structure from semantic segmentation), a ventral pathway (colour and texture from colourimetry), and a visual-cortex pathway (ResNet50 visual features). \textbf{Stage 4 (Inference)} deploys explainable AI and geospatial metrics for inference modelling, integrating multi-model comparison across semantic segmentation, land cover, and urban form variables. \textbf{Stage 5 (Result Discussion)} examines spatial distribution patterns in both horizontal and vertical dimensions, employing Moran's I and hotspot analysis, Shapley Additive exPlanations (SHAP) feature attribution for variable importance, and exploration of non-linear impacts between geospatial metrics and perceptions.

Within this framework, two of the stages involve conceptually distinct modelling steps that are easily conflated but serve different purposes, summarised in Table \ref{tab:pred_vs_inf}. The \textit{prediction model} (Stage 3) learns to map the visual content of a window view image to its perception scores, enabling us to extrapolate from the 499 surveyed images to all 12,334 citywide WVIs and map their spatial distribution. The \textit{inference model} (Stage 4) instead takes the resulting citywide perception scores as targets and regresses them on built-environment variables, in order to explain \textit{which} environmental factors drive each perception and how. These two models are detailed in the corresponding subsections below.

\begin{table}[htb]
    \centering
    \caption{Comparison between the perception prediction model and the perception inference model.}
    \label{tab:pred_vs_inf}
    \footnotesize
    \begin{tabularx}{\textwidth}{p{0.15\textwidth} X}
        \toprule
        \multicolumn{2}{l}{\textbf{Prediction model (Stage 3)}} \\
        \cmidrule(lr){1-2}
        \quad Goal & Predict perception from image content to enable citywide mapping \\
        \quad Data & 499 surveyed WVIs (27,477 pairwise comparisons from 304 participants) \\
        \quad Inputs & Semantic segmentation, ResNet50 features, and colourimetry \\
        \quad Method & Hybrid neural network (dorsal + ventral pathways) \\
        \quad Split & Five-fold spatial block cross-validation (60/20/20 within each fold) \\
        \quad Output & Citywide perception scores and spatial distribution maps \\
        \addlinespace
        \multicolumn{2}{l}{\textbf{Inference model (Stage 4)}} \\
        \cmidrule(lr){1-2}
        \quad Goal & Explain which built-environment factors drive perception and how \\
        \quad Data & All 12,334 WVIs with citywide scores from the prediction model \\
        \quad Inputs & Window-view composition, land cover, and urban form \\
        \quad Method & Machine-learning models with SHAP interpretation \\
        \quad Split & Fixed 70\%:30\% train--test split \\
        \quad Output & Variable importance and non-linear effect curves \\
        \bottomrule
    \end{tabularx}
\end{table}

\begin{figure}[ht]
    \centering
    \includegraphics[width=1\textwidth]{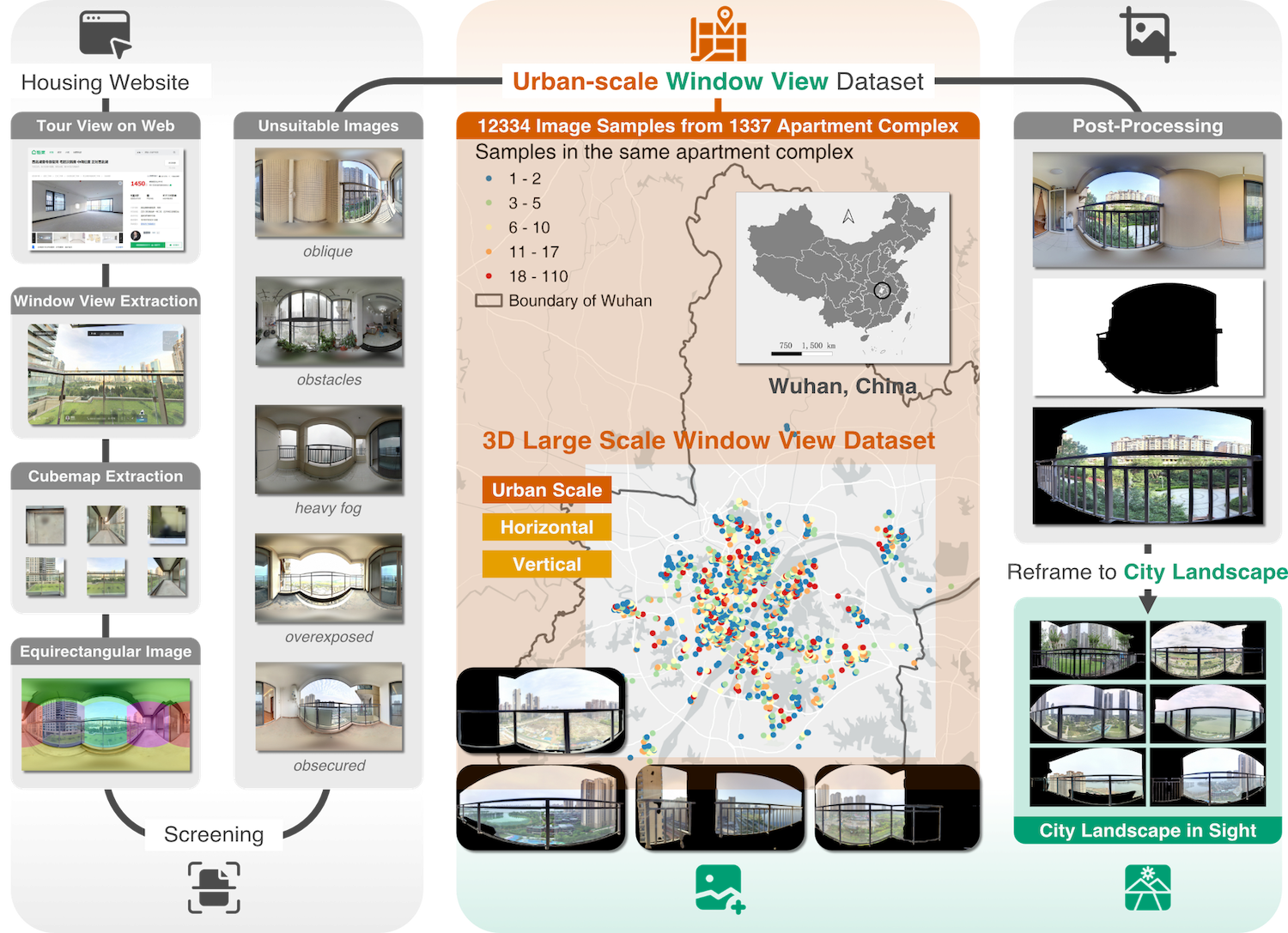}
    \caption{Data acquisition and processing workflow: (left) extraction process from housing website tour views to equirectangular images; (middle-left) examples of unsuitable images filtered during screening; (center) spatial distribution of 12,334 window view image samples from 1,377 apartment complexes across Wuhan, showing sample density per complex and coverage of both urban core and suburban areas; (right) post-processing and reframing to city landscape perspective. Source of the basemap: ESRI.}
    \label{fig_location}
\end{figure}

We selected Wuhan, a major metropolitan city in central China, as our study area due to its well-suited built environment for window view perception research (Figure \ref{fig_location}). The city exhibits high population density, with residential buildings of diverse heights and densities, and numerous rivers and lakes that contribute to a rich visual diversity of urban and natural landscapes \citep{peng2025measuring}. Spatial heterogeneity in both built form and natural resources in city landscapes provides a varied set of window view scenarios for our study, encompassing the horizontal and vertical distributions of viewpoints.

The foundation of our analysis relies on WVIs extracted from real estate listing platforms --- an emerging yet underutilised data source in urban research. These property listing platforms typically operate at regional or national scales and contain extensive textual-visual documentation of properties, enabling large-scale analyses across diverse urban research domains, such as extracting building amenity information~\citep{chen2022mining}, exploring indoor decoration patterns~\citep{liu2019inside}, discovering informal housing markets~\citep{harten2021real}, conducting comprehensive market analyses \citep{boeing2017new}, and analysing textual description patterns in property markets~\citep{lee2023online}. Besides textual information and other various attributes, such platforms contain interior imagery of properties (e.g.\ photos of bedrooms and floor plans), but the subset of these images showing window views has not been sufficiently leveraged in urban studies~\citep{Koch_2019}.

For this study, we collected window view images from Lianjia, one of China's largest real estate platforms offering comprehensive property listings nationwide. The platform's visual tour service provides 360-degree panoramic views from balconies for many listed properties, offering real perspectives of residential window views across diverse locations and building heights. We developed a web-scraping script to systematically extract cubemap data from balcony viewpoints of 34,091 residential properties across Wuhan. These cubemaps were subsequently converted into equirectangular panoramic images to facilitate analysis and perception assessment (Figure \ref{fig_location}, left panel). All WVIs in this dataset originate exclusively from living room balcony viewpoints. The platform assigns room-type labels to each panoramic capture, in which the living room balcony is consistently tagged as ``Balcony A'', while balconies associated with other spaces (e.g.\ bedrooms) are tagged as ``Balcony B'', ``Balcony C'', and so forth; our scraping script collected only images labelled as ``Balcony A'', and the manual screening described below further removed a small number of edge cases in which the living room lacked a balcony but a bedroom balcony carried the label, so that all retained images consistently represent living room balcony views. In Chinese high-density residential buildings, living room balconies are outward-facing semi-open communal spaces that serve as the primary vantage points from which residents experience the surrounding city landscape; restricting the dataset to this single room type minimises privacy-related confounds and ensures functional consistency across samples.

Following initial WVI data crowdsourcing, we implemented a WVI quality control process to ensure the reliability of these images, as illustrated in Figure \ref{fig_location}. Here, ``quality'' refers to the technical fidelity of the image as a usable record of the window view (e.g.\ capture angle, exposure, and the absence of obstructions), and not to the perceived quality of the view itself, which is precisely what the perception survey later measures. Manual screening was conducted to review, identify, and remove unsuitable images exhibiting such technical quality issues as oblique window view angles, physical obstructions (e.g., furniture, indoor objects), heavy fog or adverse weather conditions outside, image overexposure or underexposure, and obscured window views (Figure \ref{fig_location}, middle-left panel). This manual screening process yielded 12,334 technically valid WVIs from 1,377 residential apartment complexes. As depicted in the distribution map in the centre of Figure \ref{fig_location}, these WVI samples demonstrate substantial spatial coverage across Wuhan's urban and suburban areas, with sample density per apartment complex ranging from 1 to 110 images, capturing vertical (different floor levels) and horizontal (different places) distribution of window views. The retained images were further post-processed and reframed to emphasise the city landscape scope visible through windows (Figure \ref{fig_location}, right panel).
 
\subsection{Online WVI perception survey}

\begin{figure}[!ht]
    \centering
    \includegraphics[width=1\textwidth]{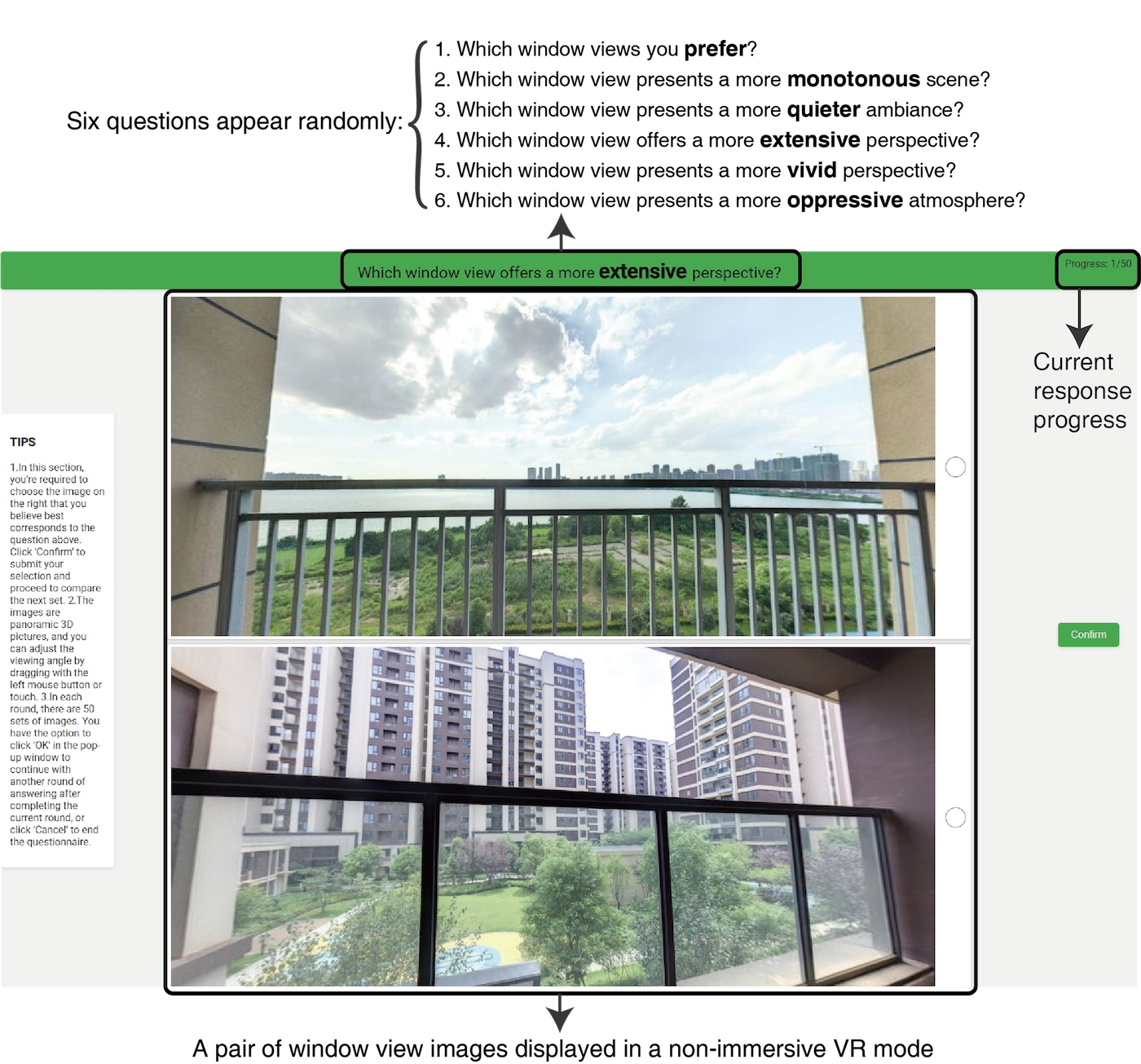}
    \caption{Web-based interface for subjective perception survey using non-immersive VR: (left) instructional tips panel; (centre) paired window view images with interactive 360-degree viewing capability and perceptual dimension question; (right) response progress indicator and confirmation button.}
    \label{fig_webside}
\end{figure}

To comprehensively assess human multidimensional subjective perception of window views, we selected six perceptual dimensions representing distinct aspects of visual-spatial experience: Prefer, Monotonous, Quiet, Extensive, Vivid, and Oppressive. The Prefer dimension serves as the primary indicator of overall window view preference, widely adopted in window view perception studies \citep{lin2022evaluation,kent2023predicting}, to capture holistic human mental aesthetic judgements. There are two dimensions characterising visual richness: Monotonous reflects perceptual uniformity and repetitiveness, while Vivid captures visual diversity and liveliness. Extensive and Oppressive dimensions address spatial qualities: Extensive measures perceived visual openness, while Oppressive assesses feelings of visual enclosure \citep{wang2024assessing,wang2024exploring}. The Quiet dimension evaluates whether participants can visually infer potential acoustic peaceful experiences from the view, representing cross-modal sensory perception \citep{chung2022study}. This six-aspect multidimensional framework enables comprehensive characterisation of window view experiences beyond aesthetic preference alone.

We developed an interactive web-based platform using Three.js to collect perceptual ratings through a non-immersive VR interface (Figure \ref{fig_webside}). The platform design emulates real estate listing interfaces familiar to participants, enhancing ecological validity. By presenting interactive panoramas through a standard web browser rather than a head-mounted display, this non-immersive design deliberately trades some sensory immersion for scalability --- enabling large-scale online recruitment --- while avoiding the simulation sickness associated with immersive VR headsets. The survey introduction explicitly informed participants that all images depicted views from residential apartment windows, establishing a residential context for evaluation. The goal was to measure participants' direct perceptual responses to the visible city landscape. The web interface presents two WVI panoramas simultaneously, displayed as interactive 360-degree images that participants can explore by dragging with mouse or touch input to adjust viewing angles (Figure \ref{fig_webside}, centre panel). For each of six perceptual dimensions, a comparative question is presented at the top of the screen (e.g., "Which window view offers a more extensive perspective?"). The survey employed a forced-choice paradigm: participants were required to select one of the two images for each comparison, with no ``indifferent'' or ``equal preference'' option available. This design follows established practice in pairwise urban perception surveys \citep{salesses2013collaborative,dubey2016deep}, ensuring every comparison yields a discriminative signal; any noise introduced by near-indifferent trials is distributed across the dataset and attenuated by the TrueSkill algorithm's Bayesian updating mechanism. Participants select the image that better matches the specified perceptual dimension by clicking the corresponding radio button, then confirm their choice to proceed to the next pair. The interface tracks response progress and provides instructional tips to ensure consistent engagement (Figure \ref{fig_webside}, left and right panels). This platform collected 43,591 pairwise comparisons from 501 participants across 499 sampled window view images.

To ensure data quality and response validity, we implemented three exclusion criteria to filter potentially inattentive or hasty responses: (1) all responses from participants whose mean response time fell below 3 seconds, indicating insufficient engagement; (2) all responses from participants who selected the same image position (left or right) in more than 80\% of trials, suggesting response bias or non-discriminative behaviour; and (3) individual comparison responses completed in less than 2 seconds, indicating insufficient time for meaningful evaluation. Applying these criteria sequentially removed 10,412, 1,143, and 4,559 responses, respectively, yielding 27,477 valid pairwise comparisons from 304 participants for subsequent analysis.

We employed the TrueSkill Bayesian ranking algorithm to convert pairwise comparison data into continuous perception scores for each image across all six dimensions. This approach models each image's latent perception level as a Gaussian distribution and iteratively updates beliefs based on comparison outcomes. On average, each image received 18.35 comparisons per dimension \(\left(\frac{27,477}{6 \times 2 \times 499}\right)\), exceeding the comparison density in Place-Pulse-1.0 (16 comparisons per image) \citep{salesses2013collaborative} and substantially surpassing Place-Pulse-2.0 (3.4 comparisons per image) \citep{dubey2016deep}. The average posterior standard deviation (\(\sigma\)) across all dimensions was less than 3, indicating high confidence in the estimated perception scores and confirming the reliability of the TrueSkill-derived rankings.

\subsection{Window view perception prediction model}

\subsubsection{Model architecture and training}

\begin{figure}[ht!]
    \centering
    \includegraphics[width=1\textwidth]{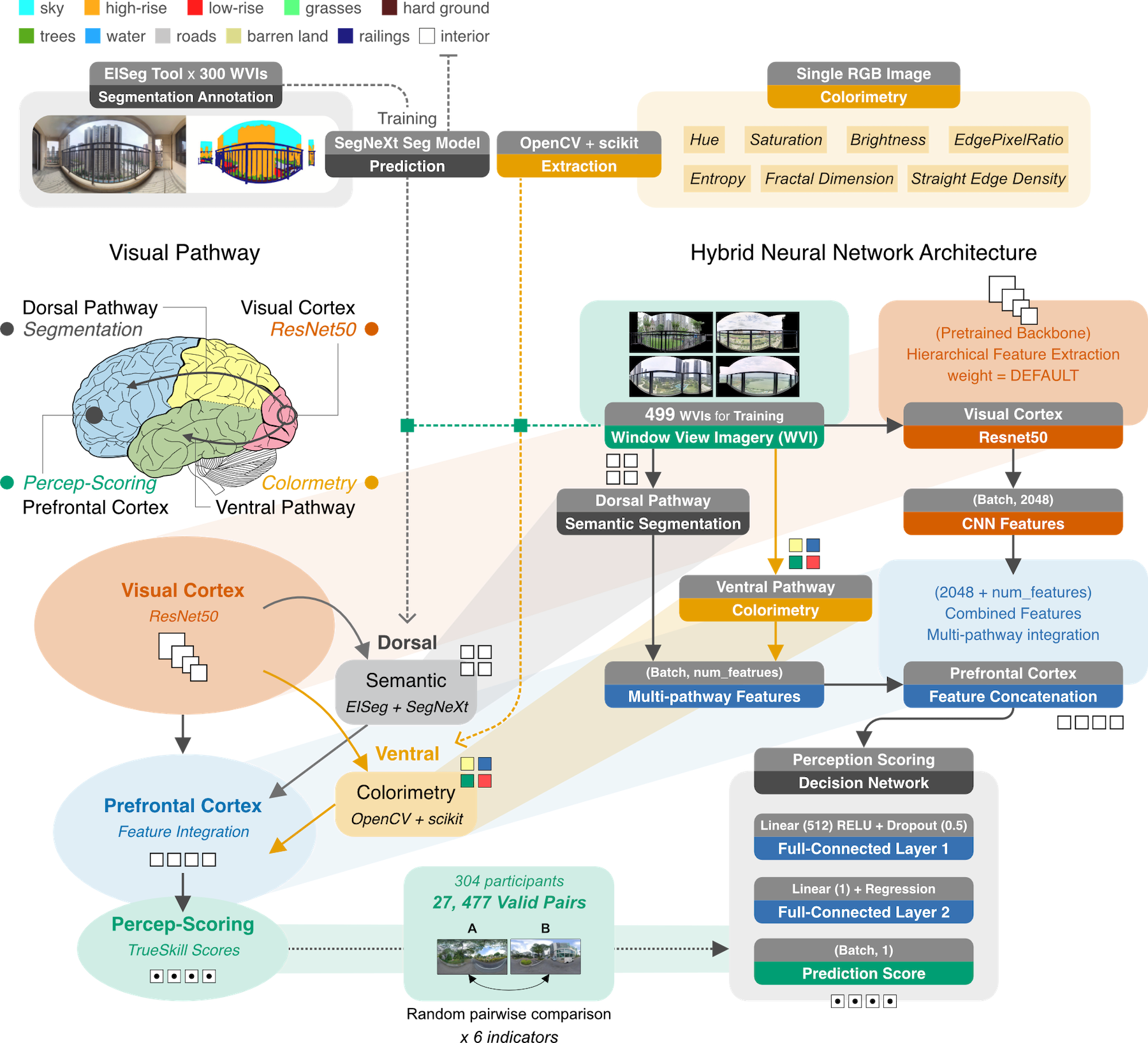}
    \caption{Hybrid neural network architecture for window view perception modelling, integrating dorsal and ventral visual pathways with multimodal feature processing.}
    \label{fig_computation}
\end{figure}

Figure \ref{fig_computation} illustrates our hybrid neural network for window view perception modelling, whose design is inspired by the visual processing pathways of the human brain. The architecture draws inspiration from neuroscientific understanding of the organisation of the visual cortex, implementing parallel processing streams that correspond to the dorsal ("where/how") and ventral ("what") pathways in human vision.

The framework integrates three complementary processing pathways: the \textbf{dorsal pathway} processes spatial and geometric information in images via semantic segmentation (EISeg + SegNeXt), extracting proportions of structural elements such as sky, buildings, and vegetation from WVIs. Specifically, we annotated 300 WVIs with the semantic labels listed in Table \ref{tab:model_variables} using EISeg \citep{hao2022eiseg}, expanded the dataset to 1,800 images through data augmentation (gamma and random gamma transformation, rotation, blurring, and noise addition), and trained a SegNeXt segmentation model \citep{guo2022segnext} on a 90\%:10\% train--validation split; the model reached a mean accuracy (mAcc) of 86.33\% and a mean Intersection over Union (mIoU) of 78.71\% before being applied to all images. The proportion of each landscape element $i$ visible through the window is defined as $WV_i = p_i / p_{total}$, where $p_i$ is the number of pixels labelled as $i$ after segmentation and $p_{total}$ is the total number of pixels in the image. The \textbf{ventral pathway} captures visual texture through colourimetry analysis (OpenCV + scikit-learn), quantifying colour distributions, contrast, and brightness characteristics of WVIs. The \textbf{visual cortex pathway} employs a pre-trained ResNet50 backbone for hierarchical feature extraction, processing raw WVIs through convolutional layers to capture high-dimensional complex visual patterns. These three pathways, together with an additional floor variable encoding the apartment's relative vertical position, constitute the model inputs summarised in Table \ref{tab:model_variables}.

These three processing streams converge in the prefrontal cortex module, where multi-pathway features are concatenated and integrated via a perception-scoring decision network. The ResNet50 backbone produces a 2048-dimensional visual embedding that is concatenated with the standardised dorsal and ventral features, and the resulting vector is passed to a decision head of two fully connected layers: the first layer (512 neurons with ReLU activation and 0.5 dropout) performs feature integration, while the second layer outputs a single continuous perception score. A separate model with this identical architecture was trained for each of the six perceptual dimensions. This multi-pathway architecture enables the model to process both low-level visual features (colour, texture) and high-level semantic content (spatial layout, object presence) in a unified framework. The models were trained to reproduce the continuous TrueSkill perception scores of the 499 sampled WVIs, which were derived from the 27,477 valid pairwise comparisons provided by 304 participants.

In designing this architecture, we made several key methodological decisions that distinguish our approach from conventional perception modelling. Rather than adopting the classification-based approaches commonly used in prior studies, we formulated the prediction task as a regression problem to preserve the granularity of continuous perception scores (1-5) derived from the TrueSkill algorithm. This design choice aligns with the continuous nature of human perceptual responses and enables the multi-pathway architecture to capture subtle perceptual nuances that would be lost in a classification method, which divides continuous scores into several discrete classes. We deliberately avoid such discretisation because partitioning the continuous scores into 2, 3, or 5 classes (e.g.\ via equal-width binning) introduces arbitrary class boundaries that discard information and complicate interpretation; we therefore report the continuous coefficient of determination $R^2$ as our primary evaluation metric throughout, rather than classification accuracy over ad hoc score bins.

The perception modelling framework was implemented with careful attention to model training strategies. Each WVI was resized to $224\times224$ pixels and normalised using the ImageNet channel statistics; during training we further applied random horizontal flipping, random rotation (up to $\pm15^{\circ}$), and colour jitter as data augmentation, while the auxiliary dorsal and ventral features were standardised using statistics fitted on the training fold only. The perception model was trained using the Adam optimiser (initial learning rate is $10^{-4}$ and weight decay is $10^{-4}$), and a step-based learning rate scheduler with a decay factor of 0.1 every 10 epochs was applied to ensure stable convergence across the different processing streams. The loss function used during training was Mean Squared Error (MSE), chosen for its compatibility with the regression modelling and for preserving the continuous nature of human perception. Training was conducted for up to 60 epochs with a batch size of 8 and early stopping (patience of 10 epochs monitored on the validation $R^2$), and the checkpoint achieving the highest validation $R^2$ was retained for evaluation. Model performance was assessed using five-fold spatial block cross-validation, in which each fold partitions the data into 60\% training, 20\% validation, and 20\% test sets while keeping geographically adjacent views within the same fold. The coefficient of determination $R^2$ served as the primary metric for both model selection and final regression model evaluation.

To clarify how the citywide perception surfaces were produced: the five-fold spatial block cross-validation provides an unbiased estimate of out-of-sample accuracy (reported in Section \ref{sec:prediction_performance}) and, in doing so, trains one model checkpoint per fold. For the subsequent citywide mapping, for each perceptual dimension we retain the single fold checkpoint that generalised best to its held-out spatial block (the highest test $R^2$) and apply this one model uniformly to all 12,334 WVIs. The reported perception maps are therefore the output of a single retained cross-validation checkpoint per dimension applied to every image.

\begin{table}[ht!]
    \centering
    \caption{Definitions and descriptive statistics of prediction model inputs.}
    \label{tab:model_variables}
    \footnotesize
    \begin{tabular}{l p{6.4cm} cccc}
        \toprule
        Variable & Definition & Mean & SD & Min & Max \\
        \midrule
        \multicolumn{6}{@{}l}{\textit{Visual cortex pathway --- raw image encoded by ResNet50}} \\
        WVI (RGB image)    & 2048-dimensional visual embedding & --- & --- & --- & --- \\
        \midrule
        \multicolumn{6}{@{}l}{\textit{Dorsal pathway --- semantic features (image-area proportion of each segmentation class)}} \\
        Sky                & Sky                                                          & 0.114 & 0.056 & 0.002 & 0.304 \\
        High rise building & Buildings with $\geq 7$ floors                               & 0.102 & 0.059 & 0.000 & 0.294 \\
        Low rise building  & Buildings with $< 7$ floors                                  & 0.013 & 0.017 & 0.000 & 0.101 \\
        Grass              & Grass                                                        & 0.007 & 0.013 & 0.000 & 0.130 \\
        Hard ground        & Community roads, sidewalks, parking and paved areas          & 0.010 & 0.011 & 0.000 & 0.083 \\
        Tree               & Trees or shrubs                                              & 0.044 & 0.044 & 0.001 & 0.438 \\
        Water              & Water bodies (rivers, lakes, etc.)                           & 0.002 & 0.007 & 0.000 & 0.097 \\
        Railing            & Railings                                                     & 0.085 & 0.030 & 0.020 & 0.208 \\
        Road               & Main avenue or railway outside the community                 & 0.003 & 0.007 & 0.000 & 0.087 \\
        Barren land        & Non-hardened barren land without cover                       & 0.002 & 0.007 & 0.000 & 0.086 \\
        Building interior  & Indoor scene and the facade of the host building in view     & 0.618 & 0.097 & 0.252 & 0.845 \\
        \midrule
        \multicolumn{6}{@{}l}{\textit{Ventral pathway --- colour and low-level visual features}} \\
        Hue\_Mean          & Mean of the HSV hue channel ($0$--$179$)                     & 58.92  & 12.17   & 19.97   & 93.45    \\
        Hue\_Std           & Standard deviation of the HSV hue channel                    & 51.06  & 5.80    & 26.64   & 65.98    \\
        Saturation\_Mean   & Mean of the HSV saturation channel ($0$--$255$)              & 40.68  & 11.19   & 12.68   & 79.75    \\
        Saturation\_Std    & Standard deviation of the HSV saturation channel             & 66.74  & 9.17    & 39.52   & 91.43    \\
        Brightness\_Mean   & Mean of the CIELab lightness ($L$) channel                   & 112.55 & 22.50   & 52.61   & 168.22   \\
        Brightness\_Std    & Standard deviation of the CIELab lightness channel           & 96.33  & 7.50    & 67.26   & 113.99   \\
        EdgePixelRatio     & Mean response of the Canny edge map (edge density)           & 42.04  & 8.87    & 14.69   & 64.59    \\
        Entropy            & Shannon entropy of the grayscale histogram (bits)            & 5.091  & 0.202   & 4.287   & 5.388    \\
        Colorfulness       & Hasler--S\"usstrunk colourfulness metric                     & 14.72  & 4.89    & 5.86    & 37.57    \\
        Contrast           & Standard deviation of grayscale intensities                  & 94.51  & 7.87    & 64.53   & 112.22   \\
        Sharpness          & Variance of the Laplacian (focus / edge sharpness)           & 3707.5 & 1089.3  & 1258.6  & 8053.2   \\
        Image\_Variance    & Variance of grayscale intensities                            & 8993.2 & 1447.1  & 4164.2  & 12592.4  \\
        \midrule
        \multicolumn{6}{@{}l}{\textit{Contextual variable}} \\
        Floor              & Relative floor-level category ($1=$ low, $2=$ middle, $3=$ high) & 1.92 & 0.78 & 1.00 & 3.00 \\
        \bottomrule
    \end{tabular}
\end{table}

\subsubsection{Spatial sampling units and cross-validation schemes}

\begin{figure}[!ht]
    \centering
    \includegraphics[width=1\textwidth]{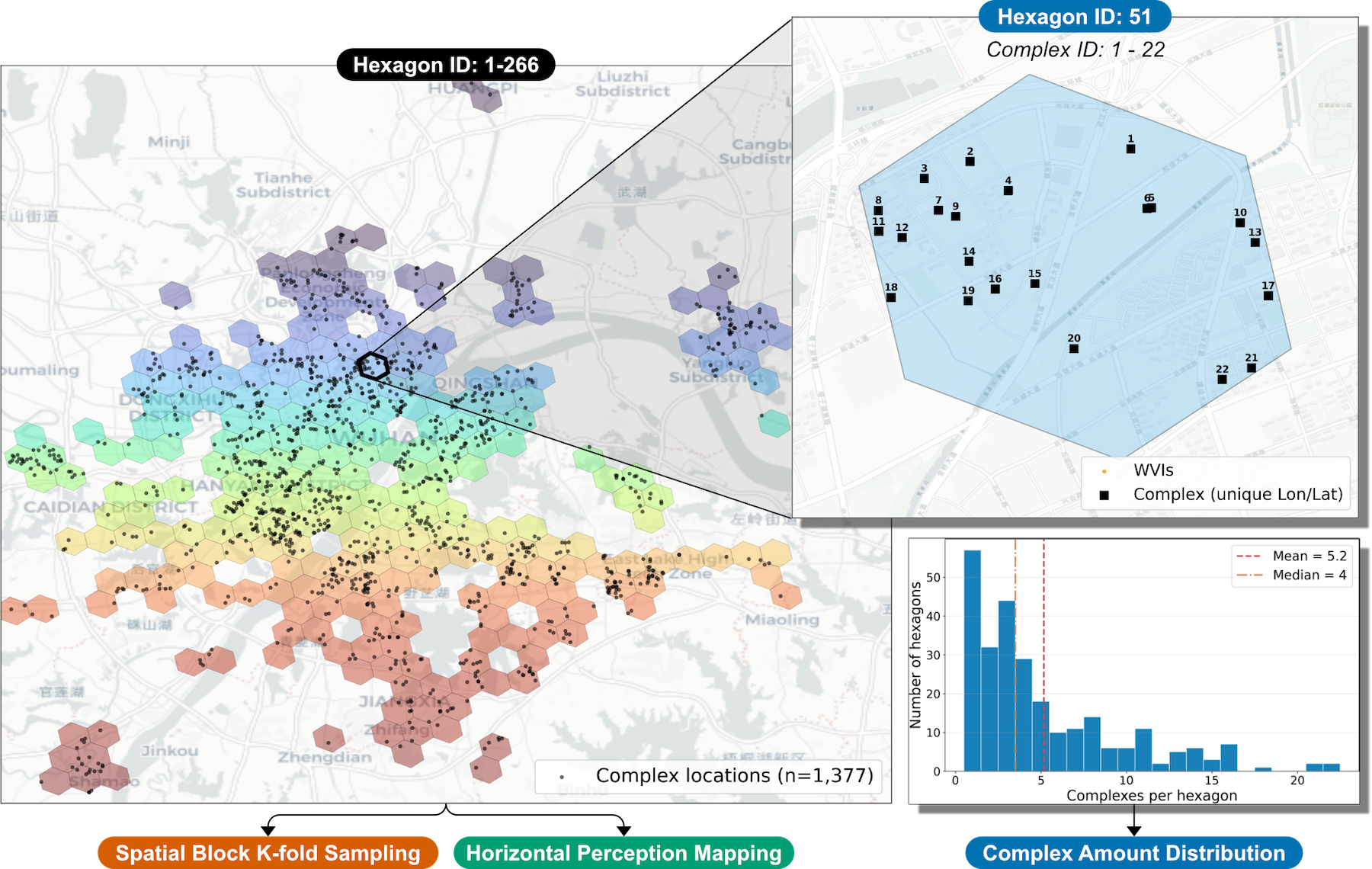}
    \caption{Nested spatial sampling units over Wuhan. (Left) the 266 H3 hexagonal cells (resolution 7) coloured by Hexagon ID, with the 1,377 complex locations as black points. (Top right) a representative hexagon (ID 51) with its 22 distinct complexes. (Bottom right) distribution of complexes per hexagon (mean 5.2, median 4, range 1--22). Basemap: CartoDB Positron.}
    \label{fig_hexagon}
\end{figure}

Window view samples from the same apartment complex share identical latitude and longitude coordinates, so we organised the 12,334 WVIs into nested spatial units for sampling and analysis. Using an H3 hexagonal grid at resolution 7, each image was assigned two hierarchical spatial identifiers. The \textit{Hexagon ID} is an integer index of the H3 cell containing the sample, numbered from 1 to 266 in geographic order (north to south, then west to east). Within each hexagon, the \textit{Complex ID} (formatted as \texttt{\{Hexagon ID\}-\{k\}}) further distinguishes each unique apartment complex, i.e.\ each distinct coordinate pair, following the same geographic ordering. In total, the dataset spans 266 hexagonal cells and 1,377 apartment complexes, with a median of 4 complexes per hexagon (range 1--22) and a median of 5 WVIs per complex (range 1--110), as illustrated in Figure \ref{fig_hexagon}. These nested identifiers define the spatial units used both for aggregating perception scores in the maps below and for the cross-validation of the prediction model.

These spatial units also underpin two complementary five-fold cross-validation schemes used to evaluate the prediction model, each trained and evaluated separately for every perception dimension with a 60/20/20 train/validation/test ratio within each fold. Both schemes are stratified on the binned perception score \emph{of the dimension being modelled} (a separate set of splits is generated per dimension), where each continuous TrueSkill score $s \in [0,5]$ is discretised into one of five bins
\[
b(s) = \min\!\left(\lfloor s \rfloor,\, 4\right) \in \{0,1,2,3,4\}.
\]
In the \textit{random stratified} scheme, the individual WVIs are split at the image level using this stratification, so that images from the same hexagon may fall into both the training and test folds. In the \textit{spatial block} scheme, whole hexagons (Hexagon IDs) are instead assigned as indivisible groups to the training, validation, and test partitions---stratified on the binned per-hexagon mean score---so that no hexagon is shared between training and test. As summarised in Table \ref{tab:kfold_schemes}, this difference is decisive: averaged over the five folds, the random stratified split shares 46.5 hexagons between the training and test sets, whereas the spatial block split shares none. The spatial block scheme therefore prevents the spatial-autocorrelation leakage that would otherwise inflate the apparent accuracy and provides a stricter, geographically honest test of generalisation; we adopt it as the primary evaluation protocol and report the random stratified results only for comparison.

\begin{table}[htb]
    \centering
    \caption{Comparison of the two five-fold cross-validation schemes. The train--test hexagon overlap is the number of H3 cells shared between the training and test sets, averaged over the five folds (range across folds in brackets).}
    \label{tab:kfold_schemes}
    \footnotesize
    \begin{tabular}{l p{3.2cm} p{3.4cm} c}
        \toprule
        Scheme & Split unit & Stratification & Train--test hexagon overlap \\
        \midrule
        Random stratified & Individual WVI (image level) & Binned image score $b(s)$ & $46.5\ [38\text{--}55]$ \\
        Spatial block & Whole hexagon (Hexagon ID group) & Binned per-hexagon mean score & $0\ [0\text{--}0]$ \\
        \bottomrule
    \end{tabular}
\end{table}

\subsubsection{Mapping the spatial distribution of perception}

For visualisation, we aggregated the point-level perception scores to the H3 hexagonal cells defined above, computing the mean value of all samples falling within each cell, so that each map summarises the local perception level while preserving the spatial structure of the city.

To quantify the degree of spatial clustering of window view perceptions, we computed the Global Moran's I statistic, which measures the overall spatial autocorrelation of a variable (Appendix, Eq.~\eqref{eq:moran}). We defined the spatial weights $w_{ij}$ as a row-standardised distance band, setting $w_{ij}=1$ when two locations lie within $1{,}000$~m of each other and $0$ otherwise. Significance was assessed with the standardised score $z=(I-\mathbb{E}[I])/\sqrt{\operatorname{Var}(I)}$ against the null hypothesis of spatial randomness; a positive and significant $I$ indicates that locations with similar perception values tend to cluster together.

While Moran's I summarises clustering globally, it does not reveal \emph{where} clusters occur. We therefore computed the local Getis-Ord $G_i^*$ statistic (Appendix, Eq.~\eqref{eq:gistar}) to locate statistically significant hot spots and cold spots, using spatial weights derived from the six nearest hexagonal neighbours of each cell (including the focal cell itself). The resulting $G_i^*$ is a z-score: large positive values identify hot spots (spatial clusters of high perception scores) and large negative values identify cold spots, with significance obtained from the standard normal distribution.

Additionally, considering that perceptions may vary between floors, we conducted a statistical analysis of perceptions based on the floor classifications provided by \href{https://lianjia.com}{lianjia.com}\footnote{%
For privacy protection, \href{https://lianjia.com}{lianjia.com} does not provide precise floor data for the apartments.}. Floor levels are categorised into three types---low floor, middle floor, and high floor---according to the following criteria:

\begin{equation*}
    \begin{cases}
    \text{Low Floor} & \text{when } F \leq 3 \text{ or } f \leq \left\lfloor \frac{F}{3} \right\rfloor \\
    \text{Middle Floor} & \text{when } \left\lfloor \frac{F}{3} \right\rfloor + 1 \leq f \leq 2 \left\lfloor \frac{F}{3} \right\rfloor \\
    \text{High Floor} & \text{when } f > 2 \left\lfloor \frac{F}{3} \right\rfloor 
    \end{cases}
\end{equation*}

where $f$ is the floor of the apartment and $F$ is the total number of floors in the building. To test whether perception scores differ across these floor categories, we applied the two-sided Mann--Whitney $U$ test pairwise between groups---a non-parametric test that compares the distributions of two independent samples without assuming normality, and is therefore well suited to the bounded, non-Gaussian perception scores.

\subsection{Window view perception inference model}

\subsubsection{Selected built environment variables influencing window view perception}

Another objective of this study is to investigate the factors influencing multidimensional window view perception. Specifically, we focus on the effects of window view composition, surrounding land use, and building forms on window view perception.

\paragraph{(1) Window view variables}\mbox{}

The first set of predictors describes the window-view composition, quantified as the proportions ($WV_i$) of the eleven landscape elements obtained with the same SegNeXt semantic segmentation model used for the prediction model's dorsal pathway (defined above; the class labels and definitions are listed in Table \ref{tab:model_variables}).

\paragraph{(2) Land cover variables}\mbox{}

We hypothesize that the land cover surrounding the window view samples may also indirectly influence people's perceptions of the window views. The Land Cover dataset, released by ESRI in 2023 with a 10-metre resolution, was clipped to obtain land cover surrounding the window view samples \citep{karra2021global}. This dataset contains nine categories: water, trees, flooded vegetation, crops, built areas, bare ground, snow/ice, clouds, and rangeland. However, since some of these categories are extremely rare or non-existent in the Wuhan area, we retained only the land cover types of water, trees, crops, built areas, bare ground, and rangeland for analysis. We created a 1 km buffer around each window view sample and calculated the proportion of selected land cover types within each buffer.

\paragraph{(3) Urban building form variables}\mbox{}

The building forms surrounding the housing samples were incorporated as potential variables influencing perceptions of window views. We analyse building density, floor area ratio, average building height, standard deviation of building height, Normalised Difference Built-up Index (NDBI), and Normalised Difference Vegetation Index (NDVI), each computed within a 1 km buffer; their definitions and descriptive statistics are summarised in Table \ref{descriptive_stats}. Building footprint data are sourced from 3D-GloBFP \citep{che2024untitled}, while NDBI and NDVI are calculated using Landsat 8 \citep{xiang2023seasonal}. 

\subsubsection{Inference modelling}

\begin{table}[htbp]
    \centering
    \caption{Definitions and descriptive statistics of the built-environment inference variables. SWIR, NIR and Red denote the shortwave-infrared, near-infrared and red band reflectance, respectively.}\label{descriptive_stats}%
    \footnotesize
    \begin{tabular}{l p{6.2cm} cccc}
        \toprule
        Variable & Definition & Mean & SD & Min & Max \\
        \midrule
        \multicolumn{6}{@{}l}{\textit{Perception variables --- TrueSkill scores rescaled to 0--5}} \\
        Prefer     & Degree of overall preference for the view & 2.553 & 1.382 & 0.000 & 5.000 \\
        Monotonous & Degree of perceived monotony              & 3.118 & 1.126 & 0.000 & 5.000 \\
        Quiet      & Degree of perceived quietness             & 2.633 & 1.187 & 0.000 & 5.000 \\
        Extensive  & Degree of perceived openness              & 2.756 & 1.295 & 0.000 & 5.000 \\
        Vivid      & Degree of perceived vividness             & 2.452 & 1.307 & 0.000 & 5.000 \\
        Oppressive & Degree of perceived oppressiveness        & 3.000 & 1.195 & 0.000 & 5.000 \\
        \midrule
        \multicolumn{6}{@{}l}{\textit{Window view variables --- image-area proportions, defined in Table \ref{tab:model_variables}}} \\
        \midrule
        \multicolumn{6}{@{}l}{\textit{Land cover variables --- area proportion within a 1\,km buffer}} \\
        LC\_Water        & Proportion of water         & 0.037 & 0.074 & 0.000 & 0.535 \\
        LC\_Trees        & Proportion of tree cover    & 0.006 & 0.022 & 0.000 & 0.482 \\
        LC\_Crops        & Proportion of cropland      & 0.006 & 0.030 & 0.000 & 0.339 \\
        LC\_Built\_Area  & Proportion of built-up area & 0.929 & 0.100 & 0.150 & 1.000 \\
        LC\_Bare\_Ground & Proportion of bare ground   & 0.000 & 0.002 & 0.000 & 0.053 \\
        LC\_Rangeland    & Proportion of rangeland     & 0.021 & 0.039 & 0.000 & 0.322 \\
        \midrule
        \multicolumn{6}{@{}l}{\textit{Building form variables --- within a 1\,km buffer}} \\
        BD     & Building density, $\sum A_{bi}/A_{\text{buffer}}$ (building base area over buffer area) & 0.135 & 0.061 & 0.001 & 0.347 \\
        FAR    & Floor area ratio, $\sum A_i/A_{\text{buffer}}$ (total floor area over buffer area)      & 0.860 & 0.437 & 0.002 & 2.654 \\
        BH     & Average building height (m)               & 14.337 & 5.621 & 0.646 & 44.586 \\
        BH\_SD & Standard deviation of building height (m) & 7.011 & 4.373 & 0.019 & 35.314 \\
        NDBI   & Normalised Difference Built-up Index, $(\text{SWIR}-\text{NIR})/(\text{SWIR}+\text{NIR})$ & -0.093 & 0.043 & -0.306 & 0.066 \\
        NDVI   & Normalised Difference Vegetation Index, $(\text{NIR}-\text{Red})/(\text{NIR}+\text{Red})$ & 0.350 & 0.074 & 0.035 & 0.703 \\
        \bottomrule
    \end{tabular}
\end{table}

Descriptive statistics for the variables used in this study are presented in Table \ref{descriptive_stats}. We first employed Spearman's correlation analysis to assess the relationships between window view perception and other variables. This method allows us to examine the degree of association among multiple dimensions of window view perception and to assess potential multicollinearity, which we further quantified using the variance inflation factor (VIF). Because the window-view and land-cover classes are compositional (each group's proportions sum to one), we dropped one reference variable from each group---the dominant window-view class (\textit{WV\_Buildinginterior}), the dominant land-cover class (\textit{LC\_Built\_Area}), and the most collinear urban-form variable (\textit{FAR})---after which all retained predictors have VIF $<10$ (see Appendix, Figure \ref{fig_vif}). The subsequent inference models are therefore fit on this reduced set of 20 predictors.

After confirming the potential multicollinearity among the variables mentioned above through correlation analysis, we selected several regression methods that are relatively insensitive to multicollinearity to fit the 20 reduced predictors. These methods included Lasso regression with an L1 penalty term, Ridge regression with an L2 penalty term, and Elastic Net regression that incorporates both L1 and L2 penalties, as well as Partial Least Squares (PLS) regression and several machine learning methods, such as Random Forest regression, Support Vector regression (SVR), and XGBoost regression. All inference models were trained on a fixed 70\%:30\% train--test split (\texttt{random\_state}~$=42$) using fixed hyperparameters rather than an exhaustive hyperparameter search, so that the comparison reflects each model family's behaviour under a common, reproducible configuration: Ridge ($\alpha=1.0$), Lasso ($\alpha=0.1$), Elastic Net ($\alpha=0.1$, L1 ratio $=0.5$), PLS ($\le 10$ components), Random Forest (100 trees), SVR (RBF kernel), and XGBoost (100 trees, learning rate $0.1$, maximum depth $6$); all remaining settings follow the \texttt{scikit-learn} and \texttt{XGBoost} defaults. After determining the best model using \( R^2 \) and Root Mean Square Error (RMSE), the SHAP method was employed to further interpret the model.

\newpage
\section{Results}

\subsection{Window view perception prediction performance}
\label{sec:prediction_performance}

\begin{table}[htb]
    \centering
    \caption{Five-fold cross-validation performance of perception prediction models for the six window view perception dimensions. Test $R^2$ and RMSE values are reported as mean $\pm$ standard deviation across folds.}
    \label{model_prediction_performance}
    \resizebox{\textwidth}{!}{%
    \begin{tabular}{lcccccc}
        \toprule
        Model & Prefer & Monotonous & Quiet & Extensive & Vivid & Oppressive \\
        \midrule
        \multicolumn{7}{@{}l}{\textit{Test $R^2$}} \\
        Ridge        & $0.662\pm0.016$ & $0.260\pm0.091$ & $0.337\pm0.158$ & $0.631\pm0.054$ & $0.388\pm0.043$ & $0.574\pm0.054$ \\
        Lasso        & $0.654\pm0.022$ & $0.257\pm0.052$ & $0.334\pm0.119$ & $0.627\pm0.038$ & $0.350\pm0.041$ & $0.557\pm0.058$ \\
        ElasticNet   & $0.667\pm0.022$ & $0.271\pm0.057$ & $0.357\pm0.122$ & $0.633\pm0.038$ & $0.385\pm0.047$ & $0.571\pm0.056$ \\
        PLS          & $0.660\pm0.013$ & $0.258\pm0.088$ & $0.334\pm0.160$ & $0.630\pm0.053$ & $0.386\pm0.044$ & $0.571\pm0.052$ \\
        RandomForest & $0.650\pm0.035$ & $0.367\pm0.025$ & $0.497\pm0.048$ & $0.660\pm0.060$ & $0.443\pm0.078$ & $0.590\pm0.051$ \\
        SVR          & $0.652\pm0.041$ & $0.364\pm0.063$ & $0.452\pm0.068$ & $0.618\pm0.046$ & $0.436\pm0.062$ & $0.579\pm0.047$ \\
        XGBoost      & $0.628\pm0.033$ & $0.267\pm0.046$ & $0.470\pm0.032$ & $0.592\pm0.073$ & $0.426\pm0.064$ & $0.568\pm0.037$ \\
        Hybrid NN & $\mathbf{0.748\pm0.022}$ & $\mathbf{0.491\pm0.079}$ & $\mathbf{0.505\pm0.061}$ & $\mathbf{0.690\pm0.045}$ & $\mathbf{0.562\pm0.081}$ & $\mathbf{0.643\pm0.097}$ \\
        \midrule
        \multicolumn{7}{@{}l}{\textit{RMSE}} \\
        Ridge        & $0.8402\pm0.0178$ & $1.2295\pm0.1254$ & $1.1760\pm0.1820$ & $0.8743\pm0.0583$ & $1.1168\pm0.1011$ & $0.9367\pm0.0357$ \\
        Lasso        & $0.8494\pm0.0220$ & $1.2348\pm0.1318$ & $1.1810\pm0.1480$ & $0.8806\pm0.0381$ & $1.1516\pm0.1161$ & $0.9553\pm0.0474$ \\
        ElasticNet   & $0.8328\pm0.0198$ & $1.2227\pm0.1246$ & $1.1598\pm0.1508$ & $0.8731\pm0.0400$ & $1.1203\pm0.1138$ & $0.9405\pm0.0433$ \\
        PLS          & $0.8430\pm0.0161$ & $1.2320\pm0.1286$ & $1.1784\pm0.1839$ & $0.8756\pm0.0575$ & $1.1185\pm0.1040$ & $0.9400\pm0.0351$ \\
        RandomForest & $0.8542\pm0.0551$ & $1.1403\pm0.1212$ & $1.0272\pm0.0845$ & $0.8384\pm0.0707$ & $1.0609\pm0.0819$ & $0.9196\pm0.0314$ \\
        SVR          & $0.8516\pm0.0564$ & $1.1392\pm0.0952$ & $1.0708\pm0.0936$ & $0.8909\pm0.0507$ & $1.0711\pm0.1099$ & $0.9315\pm0.0335$ \\
        XGBoost      & $0.8813\pm0.0434$ & $1.2254\pm0.1153$ & $1.0537\pm0.0596$ & $0.9184\pm0.0768$ & $1.0786\pm0.0914$ & $0.9451\pm0.0185$ \\
        Hybrid NN & $\mathbf{0.7242\pm0.0177}$ & $\mathbf{1.0254\pm0.1713}$ & $\mathbf{1.0164\pm0.0586}$ & $\mathbf{0.8024\pm0.0567}$ & $\mathbf{0.9374\pm0.0624}$ & $\mathbf{0.8507\pm0.0830}$ \\
        \bottomrule
    \end{tabular}%
    }
\end{table}

Table \ref{model_prediction_performance} presents the five-fold cross-validation performance of eight regression models across the six perception indicators, reporting both test $R^2$ and RMSE as mean $\pm$ standard deviation over folds. The hybrid neural network, which integrates visual features through its dorsal and ventral pathways, achieved the best performance on every dimension, attaining the highest $R^2$ and the lowest RMSE throughout. Its mean test $R^2$ values ranged from 0.491 (Monotonous) to 0.748 (Prefer), consistently outperforming the linear baselines (Ridge, Lasso, ElasticNet, PLS) and the machine-learning baselines (Random Forest, SVR, XGBoost). The improvement was most pronounced for the more perceptually variable dimensions such as Monotonous and Vivid, where the visual pathway raised $R^2$ by roughly 0.12 over the strongest baseline. Compared to other perception models applied to streetscape perception prediction tasks, our models achieved comparable performance \citep{ogawa2024evaluating}.

Among all perception dimensions, the hybrid model for Prefer and Extensive exhibited superior performance ($R^2$ of 0.748 and 0.690), indicating robust consistency in people's understanding of spatial extensiveness and their preferences for urban views \citep{yang2023role, liang2024evaluating, ogawa2024evaluating}. The prediction performance for Monotonous and Quiet was relatively lower ($R^2$ of 0.491 and 0.505), reflecting greater variability in people's perception of these dimensions when evaluating urban environments through visual imagery. Nevertheless, for Quiet the hybrid model still reduced RMSE to 1.0164 and clearly surpassed all baselines, demonstrating that models can effectively assess human auditory perceptions from visual cues in WVIs.

\begin{figure}[!ht]
    \centering
    \includegraphics[width=1\textwidth]{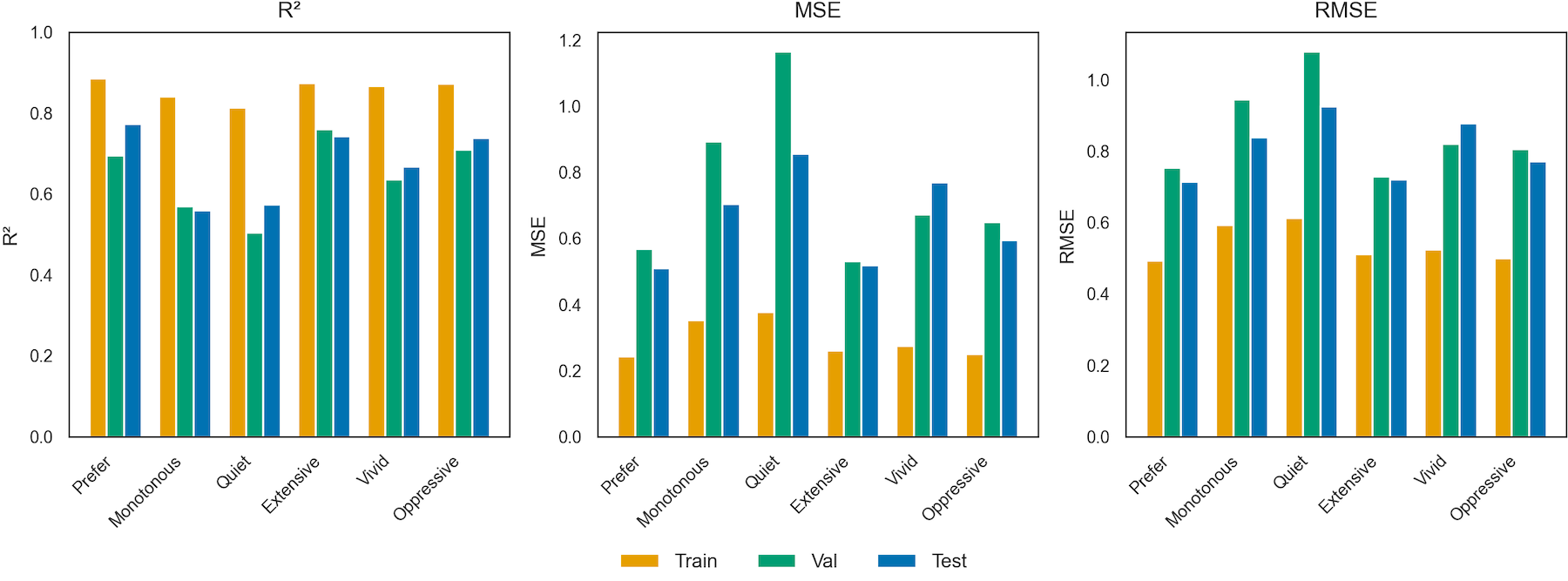}
    \caption{Per-dimension performance of the best hybrid NN checkpoint (selected on validation $R^2$) on the training, validation, and test splits: $R^2$ (left), MSE (middle), and RMSE (right).}
    \label{fig_model_prediction_best}
\end{figure}

Figure \ref{fig_model_prediction_best} reports the per-dimension performance of the best hybrid NN checkpoint across the training, validation, and test splits. On the held-out test set, the model achieved $R^2$ values ranging from 0.560 (Monotonous) to 0.773 (Prefer), with Prefer, Extensive, and Oppressive performing best ($R^2$ of 0.773, 0.743, and 0.739) and correspondingly low RMSE (0.71--0.77). The gap between the training $R^2$ (0.81--0.89) and the validation/test values indicates mild overfitting that is well controlled by early stopping, while the close agreement between the validation and test metrics confirms that the selected checkpoint generalises consistently to unseen window views. The complete per-split metrics ($R^2$, MSE, and RMSE) for all dimensions are reported in Appendix Table \ref{tab:appendix_best_metrics}.

\begin{figure}[!ht]
    \centering
    \includegraphics[width=1\textwidth]{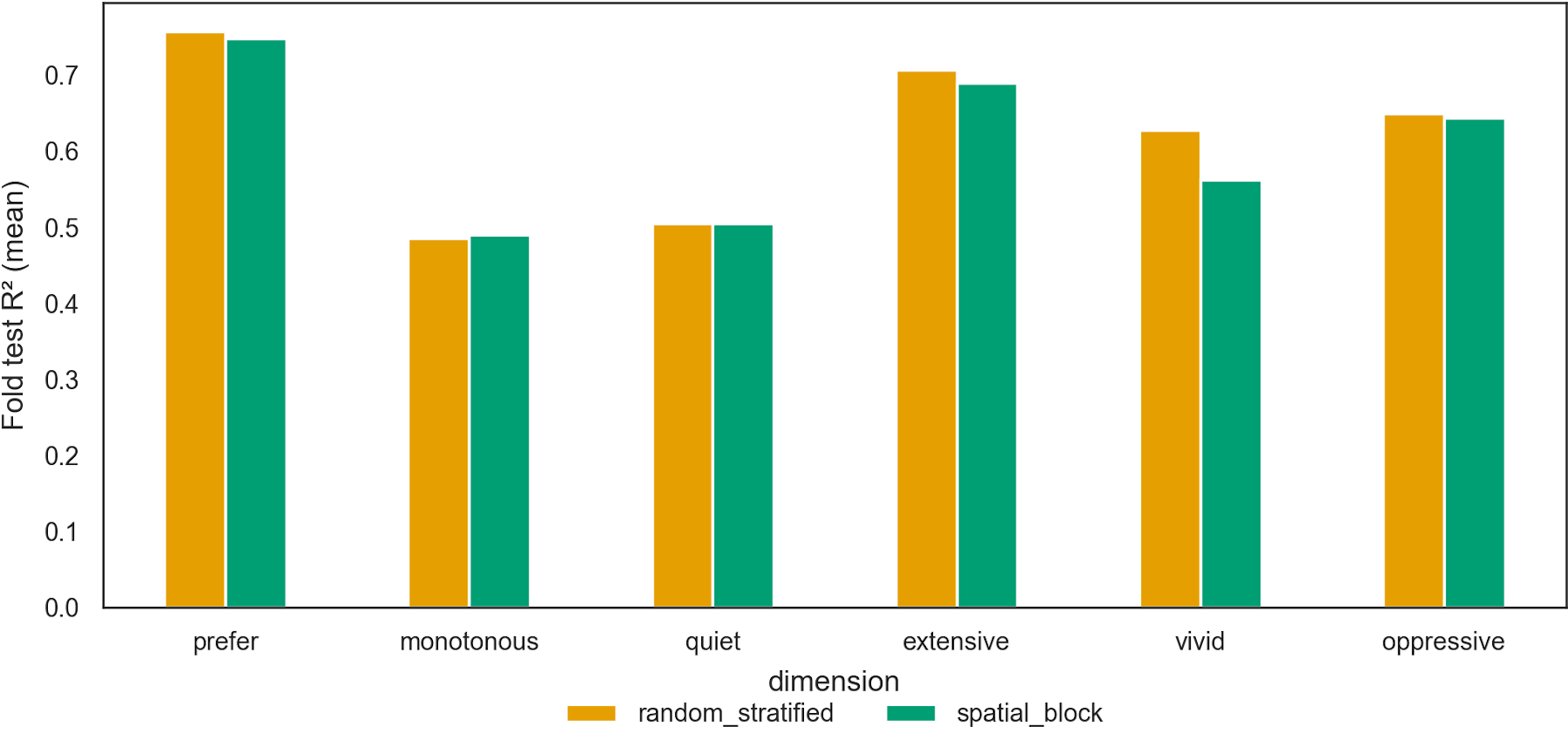}
    \caption{Comparison of the full hybrid NN performance (mean fold test $R^2$) under random stratified versus spatial block K-fold cross-validation across the six perception dimensions.}
    \label{fig_kfold_comparison}
\end{figure}

We adopted spatial block K-fold cross-validation, in which entire hexagonal blocks are held out so that the test views are geographically separated from those used for training. Figure \ref{fig_kfold_comparison} compares the mean fold test $R^2$ obtained under random stratified and spatial block splits for the six perception dimensions. The two cross-validation schemes yield broadly consistent results, confirming that the model captures genuine view--perception relationships rather than merely exploiting spatial autocorrelation. For most dimensions the spatial block $R^2$ is only marginally lower than the random stratified value (e.g.\ Prefer drops from 0.758 to 0.748 and Oppressive from 0.649 to 0.643), while Monotonous and Quiet are essentially unchanged. The largest gap appears for Vivid (0.628 versus 0.562), indicating that vividness predictions benefit the most from local visual similarity and are therefore the most affected when nearby views are withheld. Given that the spatial scheme provides the more rigorous test of geographic transferability, we report the spatial block results as our primary performance throughout; the complete per-dimension values for both schemes are listed in Appendix Table \ref{tab:appendix_kfold_comparison}.

\subsection{Ablation of visual feature pathways}

\begin{figure}[!ht]
    \centering
    \includegraphics[width=1\textwidth]{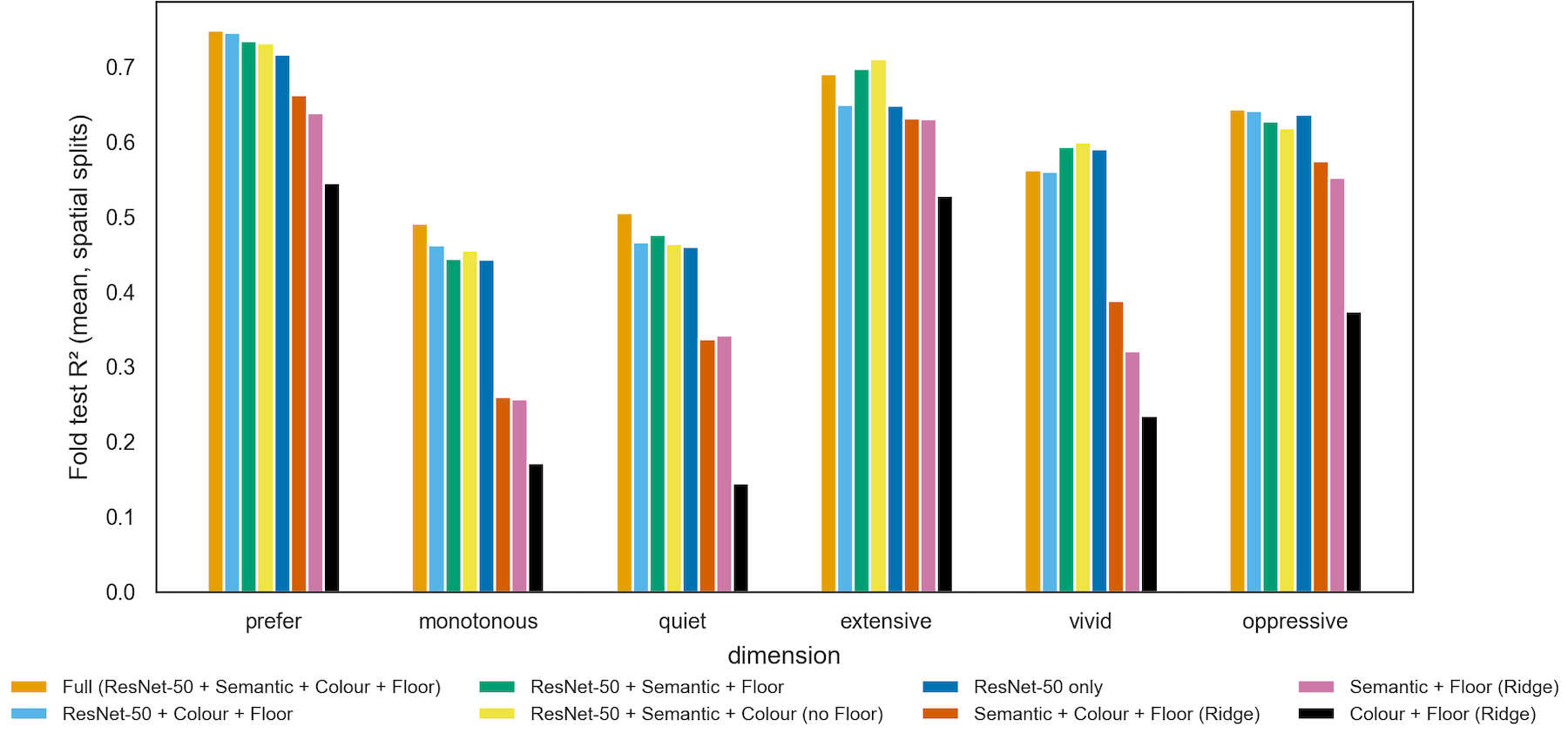}
    \caption{Pathway ablation under spatial block K-fold cross-validation. Bars show the mean fold test $R^2$ for each feature configuration across the six perception dimensions.}
    \label{fig_ablation}
\end{figure}

To quantify the contribution of each input pathway, we conducted an ablation study under the same spatial block K-fold protocol, progressively combining the ResNet50 image pathway with the semantic segmentation features, the colour features, and the floor-level variable. Figure \ref{fig_ablation} and Table \ref{tab:ablation} report the resulting fold test $R^2$ for each configuration.

\begin{table}[htb]
    \centering
    \caption{Pathway ablation results (mean fold test $R^2$, spatial block K-fold). Configurations without the ResNet50 image pathway are fitted with Ridge regression on the tabular features. The best configuration in each dimension is shown in bold.}
    \label{tab:ablation}
    \resizebox{\textwidth}{!}{%
    \begin{tabular}{lcccccc}
        \toprule
        Configuration & Prefer & Monotonous & Quiet & Extensive & Vivid & Oppressive \\
        \midrule
        ResNet50 only                            & 0.716 & 0.443 & 0.460 & 0.648 & 0.590 & 0.636 \\
        Semantic + Floor (Ridge)                 & 0.638 & 0.257 & 0.342 & 0.630 & 0.321 & 0.552 \\
        Colour + Floor (Ridge)                   & 0.545 & 0.171 & 0.145 & 0.528 & 0.234 & 0.374 \\
        Semantic + Colour + Floor (Ridge)        & 0.662 & 0.260 & 0.337 & 0.631 & 0.388 & 0.574 \\
        ResNet50 + Semantic + Floor              & 0.734 & 0.444 & 0.476 & 0.697 & 0.592 & 0.627 \\
        ResNet50 + Colour + Floor                & 0.745 & 0.462 & 0.466 & 0.649 & 0.560 & 0.641 \\
        ResNet50 + Semantic + Colour (no Floor)  & 0.731 & 0.455 & 0.464 & \textbf{0.710} & \textbf{0.599} & 0.617 \\
        Full (ResNet50 + Semantic + Colour + Floor) & \textbf{0.748} & \textbf{0.491} & \textbf{0.505} & 0.690 & 0.562 & \textbf{0.643} \\
        \bottomrule
    \end{tabular}%
    }
\end{table}

The ablation reveals that the ResNet50 image pathway is by far the dominant source of predictive signal: on its own it already attains a test $R^2$ of 0.716 for Prefer and 0.648 for Extensive, whereas the tabular pathways without the image branch are substantially weaker, with the colour-based configuration performing worst across all dimensions (e.g.\ $R^2$ of only 0.145 for Quiet and 0.171 for Monotonous). Augmenting the image pathway with semantic and colour features yields consistent gains, and the full model that additionally incorporates the floor-level variable achieves the best performance on four of the six dimensions (Prefer, Monotonous, Quiet, and Oppressive). The remaining two dimensions, Extensive and Vivid, are marginally better predicted without the floor variable ($R^2$ of 0.710 and 0.599 versus 0.690 and 0.562), suggesting that vertical position contributes little to these perceptions and may introduce minor noise. Overall, the complementary fusion of visual and environmental features in the full hybrid NN provides the most balanced and robust performance across all perception dimensions.

\subsection{Three types of window view perception emerged}

\begin{figure}[!ht]
    \centering
    \includegraphics[width=1\textwidth]{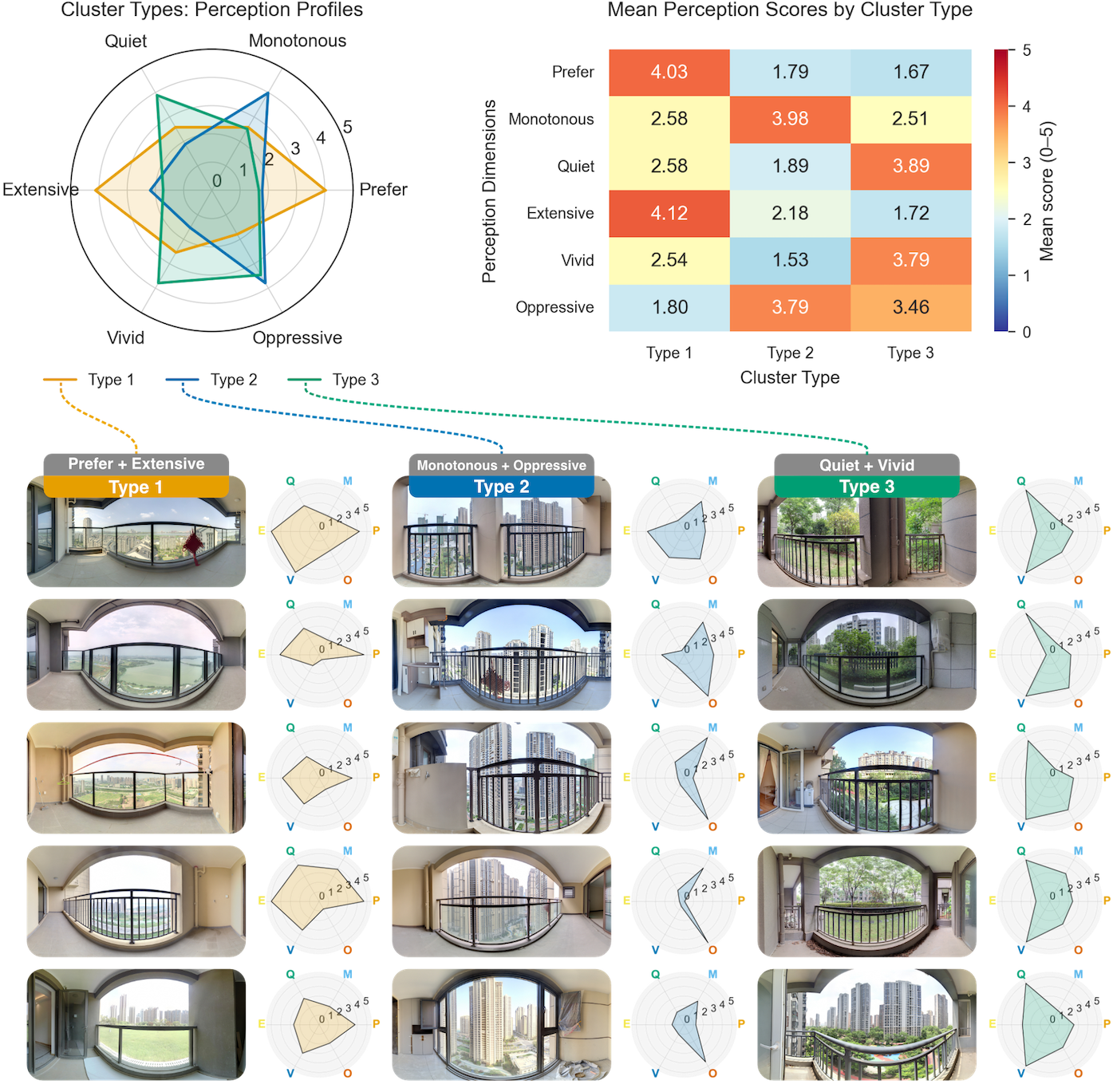}
    \caption{Three types of window view perception emerged from urban-scale window view perception prediction.}
    \label{fig_perception}
\end{figure}

Using, for each dimension, the best-performing spatial cross-validation fold checkpoint (selected on test $R^2$; Section \ref{sec:prediction_performance}), trained on the surveyed 499 WVIs, we applied the six perception prediction models to all 12,334 images for urban-scale window view perception analysis. Through clustering analysis based on the predicted perception scores, three distinct types of window view perception emerged, as illustrated in Figure \ref{fig_perception}.

The radar chart in the upper left shows the perception profiles for each cluster type, while the heatmap on the upper right displays the mean perception scores across the six dimensions. \textbf{Type 1 (Prefer + Extensive)} is characterised by high scores in Prefer (4.03) and Extensive (4.12), representing window views with expansive, unobstructed panoramas that people find most appealing. These views typically feature wide-open landscapes, distant horizons, and varied architectural elements. \textbf{Type 2 (Monotonous + Oppressive)} exhibits elevated scores in Monotonous (3.98) and Oppressive (3.79), representing constrained window views dominated by repetitive facades of many close buildings, thus with limited openness and less greenery. \textbf{Type 3 (Quiet + Vivid)} shows higher scores in Quiet (3.89) and Vivid (3.79), representing human-scale window views of more green spaces and low-rise environments that make people feel peaceful and tranquil.

The representative WVIs for each type validate the classification: Type 1 showcases viewpoints where people sweep urban vistas and natural landscapes, with the most open and extensive window views; Type 2 displays dense, oppressive urban environments with repetitive built forms; and Type 3 presents verdant, human-scale low-rise environments with abundant vegetation. These distinct perception types demonstrate the model's effectiveness in capturing the multidimensional nature of human window view perception and provide a framework for understanding urban living quality from residents' perspectives.

\subsection{Horizontal-vertical distribution of urban-scale window view perception}

To understand the spatial patterns of window view perception across the entire city, we conducted a comprehensive spatial analysis examining both horizontal distribution and vertical variation of window view perception. The study reveals significant spatial autocorrelation and clustering patterns that reflect the underlying built environment characteristics of Wuhan city.

\begin{table}[htb]
    \centering
    \caption{Global Moran's I of window-view perceptions, computed for all floors combined (one mean value per apartment-complex location) and separately for each floor level. Significance: $^{***}p<0.001$, $^{**}p<0.01$, $^{*}p<0.05$; values without a marker are not significant.}
    \label{tab_moran}
    \resizebox{\textwidth}{!}{%
    \begin{tabular}{l cc cc cc cc}
        \toprule
        & \multicolumn{2}{c}{All floors ($n=1{,}377$)} & \multicolumn{2}{c}{Low floor ($n=984$)} & \multicolumn{2}{c}{Middle floor ($n=1{,}082$)} & \multicolumn{2}{c}{High floor ($n=1{,}039$)} \\
        \cmidrule(lr){2-3}\cmidrule(lr){4-5}\cmidrule(lr){6-7}\cmidrule(lr){8-9}
        Perceptions & Moran's I & z & Moran's I & z & Moran's I & z & Moran's I & z \\
        \midrule
        Prefer      & $0.066^{**}$  & 3.245 & $0.096^{***}$ & 3.573 & $0.025$       & 0.998 & $0.050$       & 1.917 \\
        Monotonous  & $0.096^{***}$ & 4.661 & $0.082^{**}$  & 3.043 & $0.068^{**}$  & 2.690 & $0.055^{*}$   & 2.100 \\
        Quiet       & $0.070^{***}$ & 3.434 & $0.040$       & 1.510 & $0.070^{**}$  & 2.746 & $0.072^{**}$  & 2.765 \\
        Extensive   & $0.051^{*}$   & 2.519 & $0.086^{**}$  & 3.187 & $0.024$       & 0.957 & $0.075^{**}$  & 2.866 \\
        Vivid       & $0.123^{***}$ & 5.967 & $0.155^{***}$ & 5.710 & $0.112^{***}$ & 4.396 & $0.078^{**}$  & 2.963 \\
        Oppressive  & $0.071^{***}$ & 3.462 & $0.090^{***}$ & 3.339 & $0.033$       & 1.316 & $0.054^{*}$   & 2.083 \\
        \bottomrule
    \end{tabular}%
    }
\end{table}

Global Moran's I analysis confirms statistically significant positive spatial autocorrelation for all six perception dimensions (Table \ref{tab_moran}, all $p<0.05$). Vivid exhibits the strongest spatial clustering (Moran's I $=0.123$, $z=5.97$), followed by Monotonous (0.096) and a closely grouped Oppressive (0.071), Quiet (0.070) and Prefer (0.066), while Extensive shows the weakest---yet still significant---clustering (0.051, $z=2.52$, $p=0.012$). The positive and significant statistics indicate that complexes with similar window-view perceptions tend to be geographically close, reflecting the spatially structured nature of the built environment in Wuhan; the relatively modest magnitudes are consistent with perceptions aggregated at the fine, complex-level resolution rather than over coarse spatial blocks.

Disaggregating the analysis by floor level reveals that spatial clustering is not uniform across building heights (Table \ref{tab_moran}, floor-level columns). Vivid again shows the strongest and most consistent autocorrelation, remaining highly significant at every level while weakening monotonically from the low (0.155) to the high floor (0.078); this gradient suggests that the perceived vividness of low-floor views is most tightly tied to the immediate local environment, whereas higher vantage points open onto broader, more heterogeneous cityscapes that dilute local similarity. Prefer, Extensive and Oppressive exhibit the highest clustering at the low floor and lose significance at the middle floor, indicating that street-level preference, openness and oppressiveness are governed by localised ground conditions that become spatially diffuse mid-rise. In contrast, Quiet displays the opposite pattern, being non-significant at the low floor (0.040, $p=0.131$) but significantly clustered at the middle and high floors, consistent with quietness being shaped by elevation-dependent factors such as distance from street-level noise. Overall, low-floor perceptions are the most spatially structured for most dimensions, while the middle floor shows the weakest and least consistent clustering.

\begin{figure}[htb]
    \centering
    \includegraphics[width=1\textwidth]{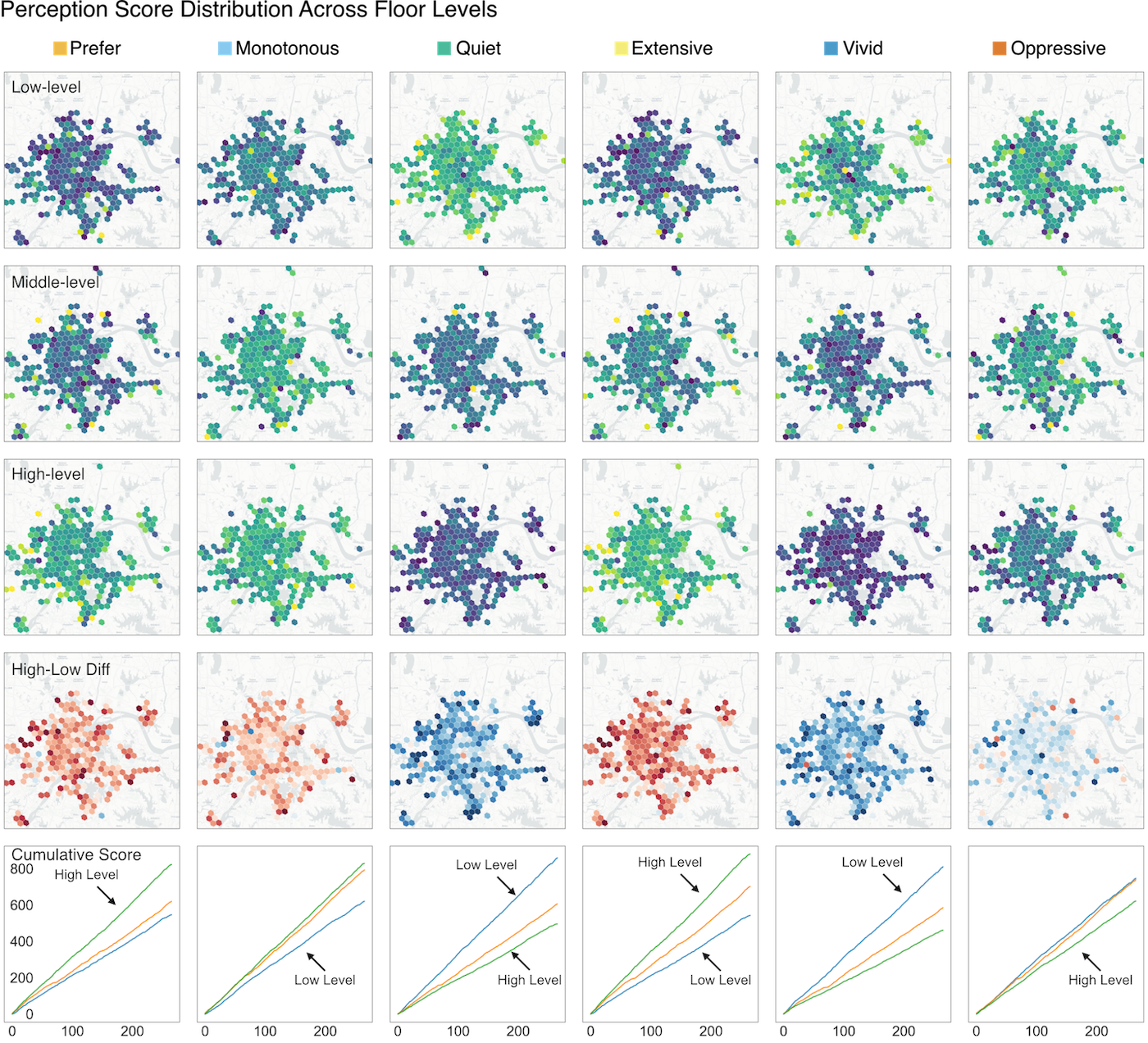}
    \caption{Urban-scale spatial distribution of six perception indicators across floor levels.}
    \label{fig_incity}
\end{figure}

Figure \ref{fig_incity} presents the urban-scale spatial distribution patterns of six perception indicators across three floor levels (low-level, middle-level, high-level) and their differences (high-low). The top three rows show perception score distributions using hexagonal grids, with colour intensity representing score magnitude. The cumulative score plots at the bottom reveal distinct patterns: higher floors generally exhibit higher scores for Prefer, Monotonous, and Extensive perceptions, while lower floors are associated with higher Quiet and Vivid perception scores. The high-low difference maps clearly illustrate these vertical gradients, with red areas indicating higher scores at upper floors and blue areas showing higher scores at lower floors.

\begin{figure}[htb]
    \centering
    \includegraphics[width=1\textwidth]{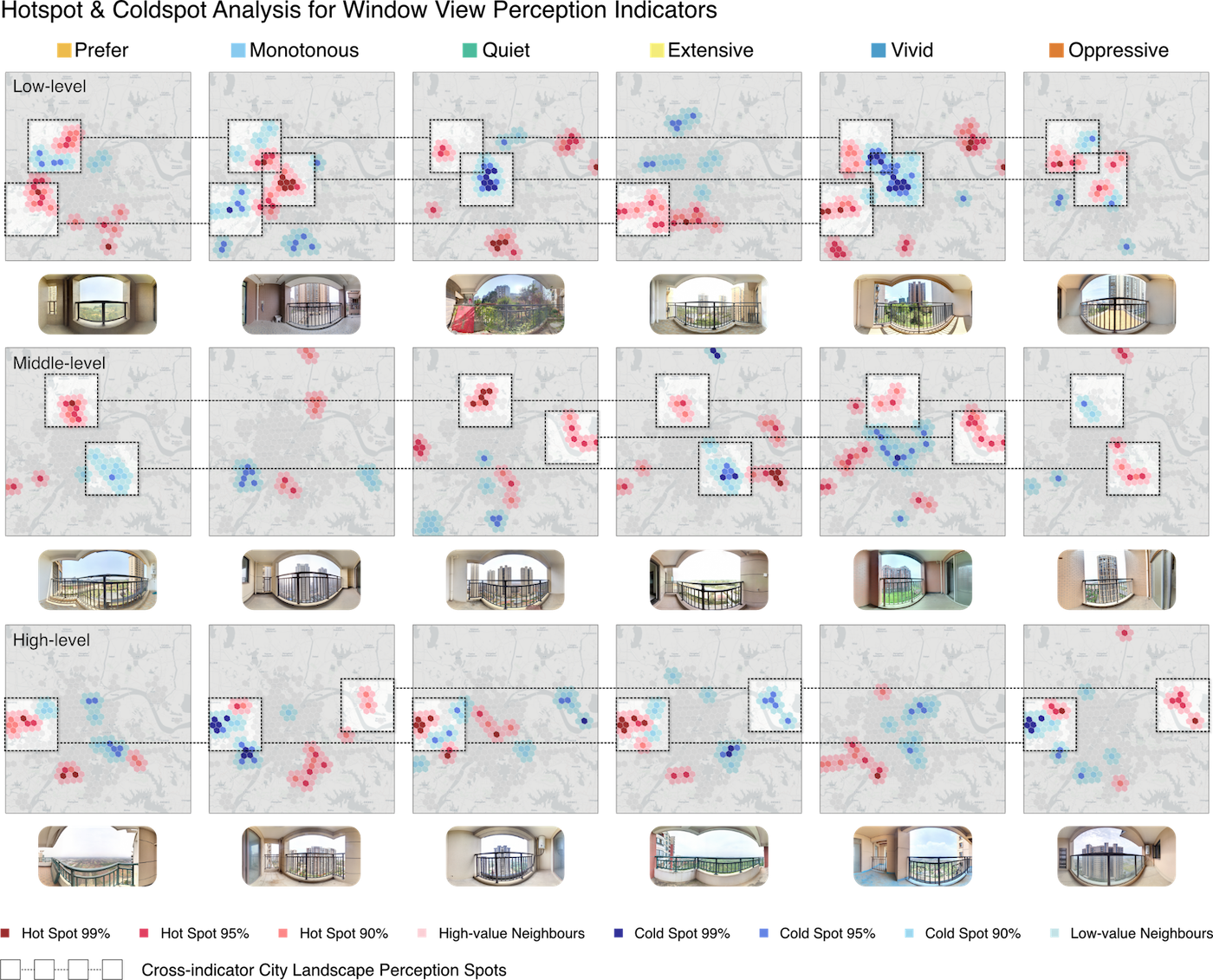}
    \caption{Hotspot and coldspot analysis of urban-scale window view perception.}
    \label{G}
\end{figure}

The hotspot and coldspot analysis (Figure \ref{G}) provides more detailed spatial insights by using local Getis-Ord Gi* statistics for low, middle, and high floor levels separately. Examining the general patterns across all floors, Prefer and Extensive show similar spatial distributions, with hotspots (red areas) typically concentrated in southwestern Wuhan and coldspots (blue areas) in rapidly developed high-rise districts, including Baishazhou, Qingshan, and Dongxihu. Conversely, Monotonous and Oppressive exhibit hotspots precisely in these high-density development areas. Quiet and Vivid display comparable patterns, with coldspots primarily in central urban areas (Hankou and Hanyang) and hotspots distributed around the city's periphery, where lower-density, greener environments prevail. However, these patterns vary considerably across low-, middle-, and high-level floors, as shown in the different rows of the analysis. The representative WVIs accompanying each map validate these spatial patterns, showing the visual characteristics of featured WVI samples in highlighted areas.

Notably, these spatial distribution patterns align well with the three types of window view perceptions identified above. Areas with \textbf{Type 1} (Prefer + Extensive) perception characteristics --- featuring expansive and appealing views --- correspond to the hotspots featuring Prefer and Extensive perception in southwestern Wuhan. \textbf{Type 2} (Monotonous + Oppressive) areas, characterised by repetitive building elements and constrained city views, match the hotspots featuring Monotonous and Oppressive perception in high-density urban districts. \textbf{Type 3} (Quiet + Vivid) areas, representing green spaces and low-rise urban environments, align with the hotspots featuring Quiet and Vivid perception around the city's periphery. This geospatial-typological correspondence demonstrates that our window view perception clustering not only captures meaningful perceptual information but also reflects the actual geographic distribution of different perceptions driven by the corresponding urban environment across Wuhan city.

\begin{figure}[htbp]
    \centering
    \includegraphics[width=1\textwidth]{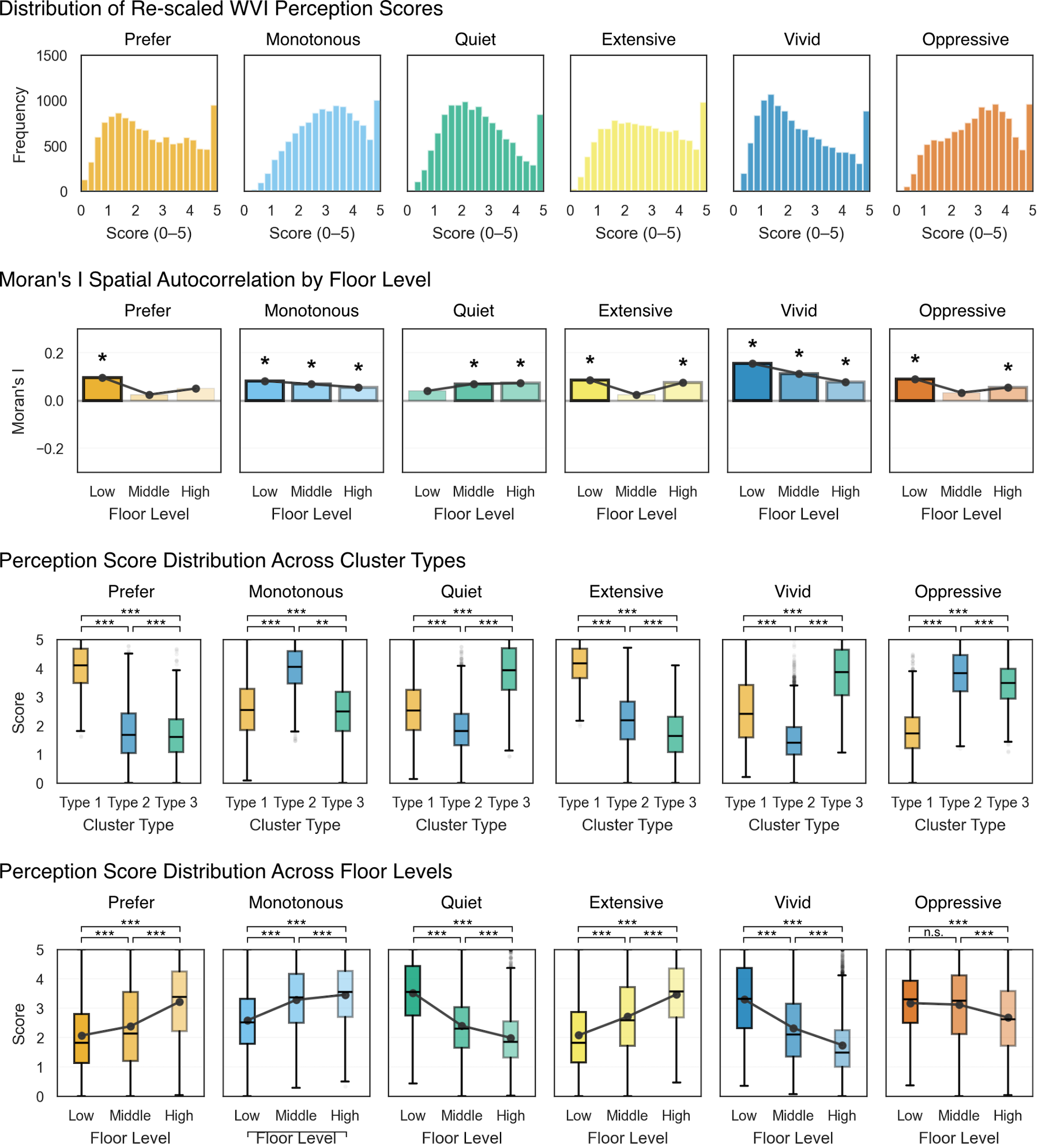}
    \caption{Differences in perception scores across floors. Statistical significance was assessed using the Mann-Whitney U test (\textbf{***}: $p$ \textless{} 0.001, \textbf{n.s.}: Not Significant).}
    \label{fig_floor}
\end{figure}

Figure \ref{fig_floor} presents a comprehensive statistical analysis of window view perception scores across multiple aspects. The distribution histograms at the top of the figure show that the positively valenced dimensions are skewed toward lower values, with Vivid, Prefer, and Quiet exhibiting the most pronounced concentration in the lower score ranges, whereas the negatively valenced Monotonous and Oppressive dimensions are centred slightly above the scale midpoint and Extensive is spread more evenly. This pattern suggests that most urban window views in Wuhan offer limited positive visual experience; the spike at the 5-point mark reflects the score rescaling, in which the top-scoring images are capped at the maximum value of 5. The cluster type analysis (third panel) validates our three-type classification of window view perception, showing significant differences ($***p < 0.001$) between types across all perception dimensions. \textbf{Type 1} consistently exhibits the highest Prefer and Extensive perception scores; \textbf{Type 2} shows higher Monotonous and Oppressive perception scores; and \textbf{Type 3} demonstrates higher Quiet and Vivid perception scores, confirming the robustness of our perception clustering.

The three floor-related analyses reveal systematic relationships between vertical floor level position and window view perception patterns. Moran's I spatial autocorrelation analysis by floor level (second panel) shows that spatial clustering patterns vary across vertical positions, with all perceptions maintaining significant autocorrelation (marked with asterisks) at most floor levels. Vivid and Monotonous perceptions demonstrate consistent geospatial clustering across all three floor levels, whereas other perceptions show varying degrees of spatial dependence by floor level. The floor-level analysis (bottom panel) reveals distinct vertical gradients with significant perceptual differences across floor levels for all perceptions except Oppressive (marked as n.s. between the low and middle floor levels). Higher floor levels are significantly associated with increased Prefer, Monotonous, and Extensive perception scores, reflecting enhanced visual access to open views and a broad city landscape. Conversely, lower floor levels exhibit significantly higher Quiet and Vivid perception scores, likely due to closer proximity to street-level vegetation and human-scale environments. The trend lines clearly illustrate these vertical gradient patterns, with Quiet perception showing a notable decline from low to high floor levels, while Prefer and Extensive perceptions show ascending trends.

\subsection{Environmental factors influencing window view perception}

\begin{figure}[htb]
    \centering
    \includegraphics[width=1\textwidth]{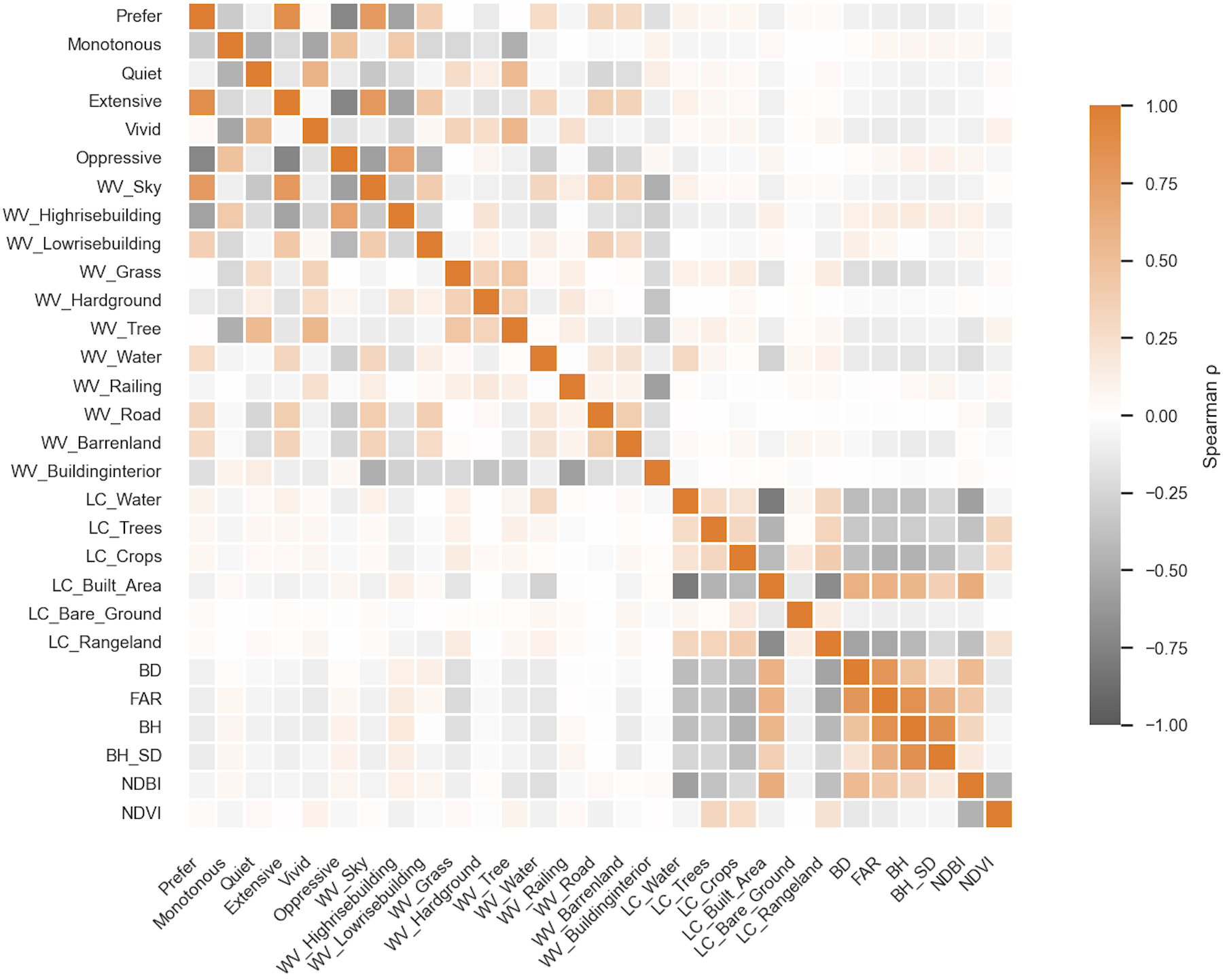}
    \caption{Correlation coefficients between six window view perception indicators and urban environment variables.}
    \label{cor}
\end{figure}

The correlation analysis among all variables in our inference modelling reveals distinct groups of correlated variables, as illustrated in Figure \ref{cor}. Highly correlated variables are predominantly located in the upper-left and lower-right corners of the figure. The upper-left cluster primarily encompasses the six window view perception dimensions along with key city landscape elements in the WVIs: \textit{WV\_Sky}, \textit{WV\_Highrise}, \textit{WV\_Lowrise}, and \textit{WV\_Tree}. Among the perception dimensions, Extensive --- a relatively positive perception --- exhibits a strong positive correlation with Prefer, whereas the negative perceptions (Monotonous and Oppressive) correlate negatively with Prefer. Quiet and Vivid show only weak correlations with Prefer but a clear positive correlation with each other, while Extensive and Oppressive display a strong negative correlation with one another. Furthermore, the strong correlations between the land cover and building form variables in the lower-right corner underscore the necessity of employing models insensitive to multicollinearity and of using VIF to filter predictors.

Performance comparisons across the candidate inference models are presented in Table \ref{comparison}. We emphasise that these inference models---which regress the built-environment variables on the perception scores to reveal which environmental factors drive perception---are separate from the hybrid neural network prediction model of Stage 3 that generates the citywide perception scores. Among the inference models, XGBoost, which combines regularisation with decision trees, achieved the highest $R^2$ values and the lowest RMSE across all six perception dimensions. The tree-based and kernel models (XGBoost, Random Forest, and SVR) consistently outperformed the linear models (Ridge, Lasso, ElasticNet, and PLS), indicating that the relationships between environmental factors and window view perception are partly non-linear. Explanatory power also varied markedly across dimensions: the models accounted for the most variance in Extensive ($R^2 = 0.80$), Prefer ($R^2 = 0.79$), and Oppressive ($R^2 = 0.74$) perceptions, but explained considerably less for Monotonous ($R^2 = 0.48$), Quiet ($R^2 = 0.55$), and Vivid ($R^2 = 0.56$) perceptions, suggesting that the latter are shaped by factors beyond the measured built-environment variables. Consequently, we employed XGBoost in conjunction with SHAP to explore how environmental factors influence multidimensional perceptions of window views.

\begin{table}[htb]
    \centering
    \caption{Comparison of \( R^2 \) and RMSE across candidate inference models. These inference models relate built-environment variables to perception scores and are distinct from the hybrid neural network used for perception prediction (Stage 3).}
    \resizebox{\textwidth}{!}{%
    \begin{tabular}{lcccccccccccc}
        \toprule
        Model & \multicolumn{2}{c}{Prefer} & \multicolumn{2}{c}{Monotonous} & \multicolumn{2}{c}{Quiet} & \multicolumn{2}{c}{Extensive} & \multicolumn{2}{c}{Vivid} & \multicolumn{2}{c}{Oppressive} \\
        \cmidrule(lr){2-3} \cmidrule(lr){4-5} \cmidrule(lr){6-7} \cmidrule(lr){8-9} \cmidrule(lr){10-11} \cmidrule(lr){12-13}
        & \( R^2 \) & RMSE & \( R^2 \) & RMSE & \( R^2 \) & RMSE & \( R^2 \) & RMSE & \( R^2 \) & RMSE & \( R^2 \) & RMSE \\
        \midrule
        Ridge         & 0.7347 & 0.7108 & 0.3859 & 0.8903 & 0.4263 & 0.8867 & 0.7514 & 0.6428 & 0.3682 & 1.0364 & 0.6833 & 0.6750 \\
        Lasso         & 0.7117 & 0.7410 & 0.3472 & 0.9179 & 0.3721 & 0.9276 & 0.7231 & 0.6784 & 0.3127 & 1.0810 & 0.6545 & 0.7049 \\
        ElasticNet    & 0.7230 & 0.7262 & 0.3685 & 0.9029 & 0.4042 & 0.9036 & 0.7387 & 0.6591 & 0.3459 & 1.0546 & 0.6682 & 0.6909 \\
        PLS           & 0.7344 & 0.7112 & 0.3853 & 0.8908 & 0.4250 & 0.8877 & 0.7514 & 0.6428 & 0.3602 & 1.0429 & 0.6842 & 0.6741 \\
        RandomForest  & 0.7726 & 0.6581 & 0.4564 & 0.8376 & 0.5254 & 0.8065 & 0.7888 & 0.5924 & 0.5211 & 0.9024 & 0.7194 & 0.6353 \\
        SVR           & 0.7722 & 0.6586 & 0.4572 & 0.8370 & 0.5178 & 0.8129 & 0.7860 & 0.5964 & 0.5096 & 0.9131 & 0.7215 & 0.6330 \\
        XGBoost       & \textbf{0.7855} & 0.6391 & \textbf{0.4788} & 0.8202 & \textbf{0.5513} & 0.7842 & \textbf{0.8027} & 0.5727 & \textbf{0.5614} & 0.8635 & \textbf{0.7379} & 0.6140 \\
        \bottomrule
    \end{tabular}%
    }
    \label{comparison}
\end{table}

Before interpreting these results, we emphasise that the XGBoost--SHAP analysis identifies statistical \textit{associations} between the built-environment variables and the predicted perception scores, not causal mechanisms. SHAP values quantify each variable's partial contribution to the model output given the observed data, but the built-environment predictors are mutually correlated (e.g.\ visible roads tend to co-occur with denser, noisier built environments), so an attribution assigned to one variable may partly reflect confounding with co-varying factors that we did not measure. The directions and magnitudes reported below should therefore be read as model-derived tendencies to be confirmed by controlled or longitudinal studies, rather than as the isolated causal effects of individual elements.

\begin{figure}[htb]
    \centering
    \includegraphics[width=1\textwidth]{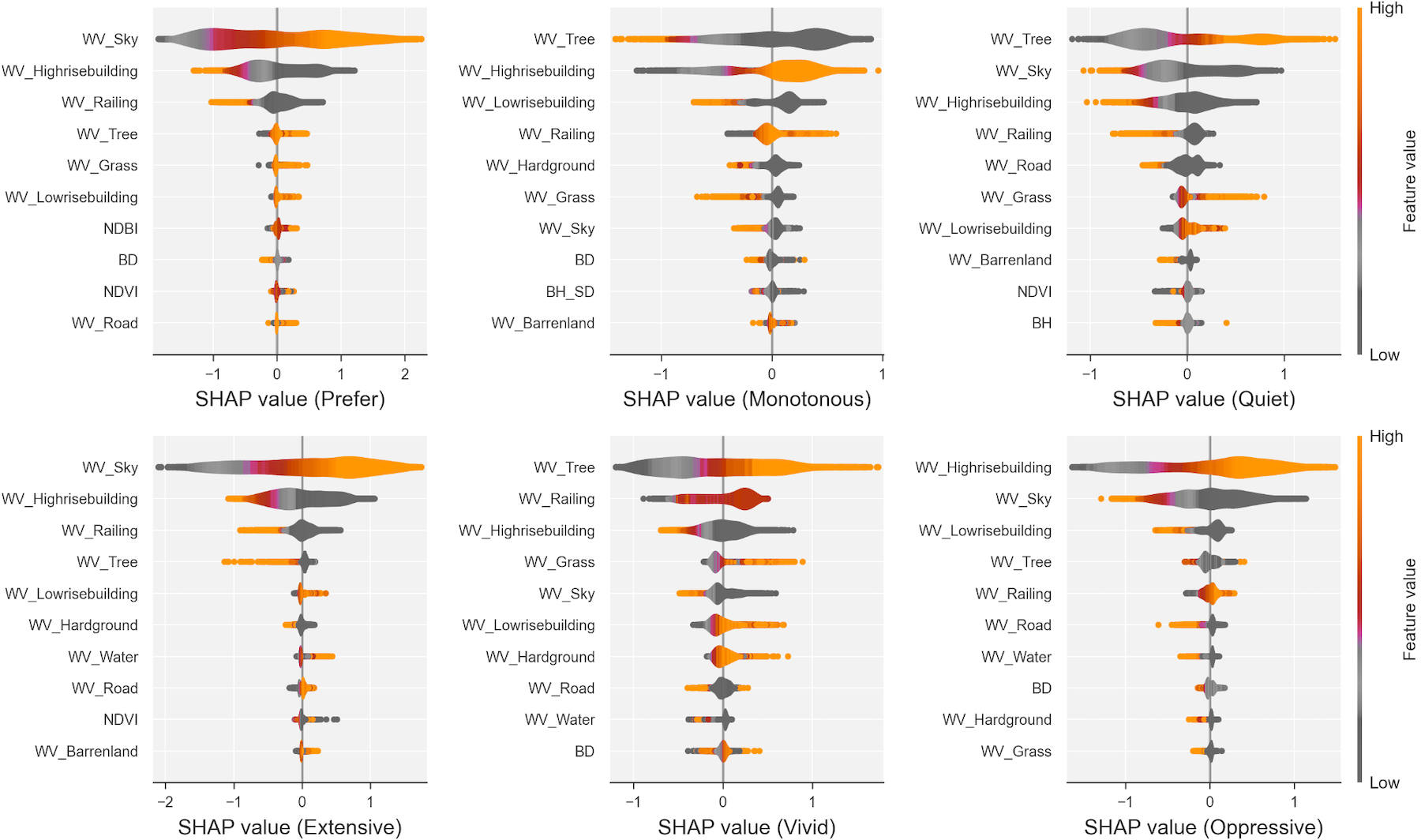}
    \caption{Variable importance analysis using XGBoost and SHAP methodology.}
    \label{SHAP}
\end{figure}

Using XGBoost and SHAP methodology, we identified the ten most influential variables in the regression models for each perception dimension, illustrated in Figure \ref{SHAP}. Across all perceptions, the top five variables consistently comprised window view semantic elements, indicating that visible city landscape features are the strongest correlates of window view perception compared to other indirect built environment indicators. Building form-related variables, such as building density (\textit{BD}) and building height standard deviation (\textit{BH\_SD}), showed minor importance for specific perceptions such as Monotonous and Vivid. In contrast, land use variables had a negligible impact on perception outcomes.

Four key window view variables --- \textit{WV\_Sky}, \textit{WV\_Highrise}, \textit{WV\_Lowrise}, and \textit{WV\_Tree} --- demonstrate substantial influence across most perception dimensions:

\begin{itemize}[label=$\bullet$]
    \item \textbf{Sky visibility} (\textit{WV\_Sky}) plays a crucial role in perceptions such as Prefer, Quiet, Extensive, and Oppressive. Increased sky visibility could enhance window views and spatial openness, helping reduce feelings of oppression. However, expanded sky views may also diminish perceived quietness, potentially due to associations with increased exposure to urban noise sources.

    \item \textbf{High-rise buildings} (\textit{WV\_Highrise}) significantly influence perceptions of Prefer, Monotonous, Extensive, Vivid, and Oppressive. A greater presence of high-rise buildings intensifies residents' feelings of oppression and monotony, suggesting that dominant vertical elements in window views could evoke psychological constraints and discomfort. Conversely, high-rise buildings tend to reduce preference, perceived openness, and vividness, indicating that extensive high-level views on high-rise buildings may compromise visual perception from nearby buildings.

    \item \textbf{Low-rise buildings} (\textit{WV\_Lowrise}) exert relatively moderate influence, contributing positively to Prefer, Extensive, and Vivid perceptions while reducing monotonous feelings. Unlike high-rise buildings, low-rise buildings do not significantly obstruct views or induce psychological pressure. Instead, their diverse architectural forms may enhance visual diversity and reduce monotony.

    \item \textbf{Vegetation} (\textit{WV\_Tree}) emerges as the most influential variable for Monotonous, Quiet, and Vivid perceptions. Vegetation can enhance feelings of tranquillity and liveliness, increase overall visual preference, and reduce Monotonous perception. However, these benefits are not unconditional: beyond a modest share of the view, dense tree cover progressively curtails perceived openness (Extensive) by obstructing distant sight lines, and very high coverage is associated with a mild rise in oppressive feelings.
\end{itemize}

\begin{figure}[H]
    \centering
    \includegraphics[width=1\textwidth]{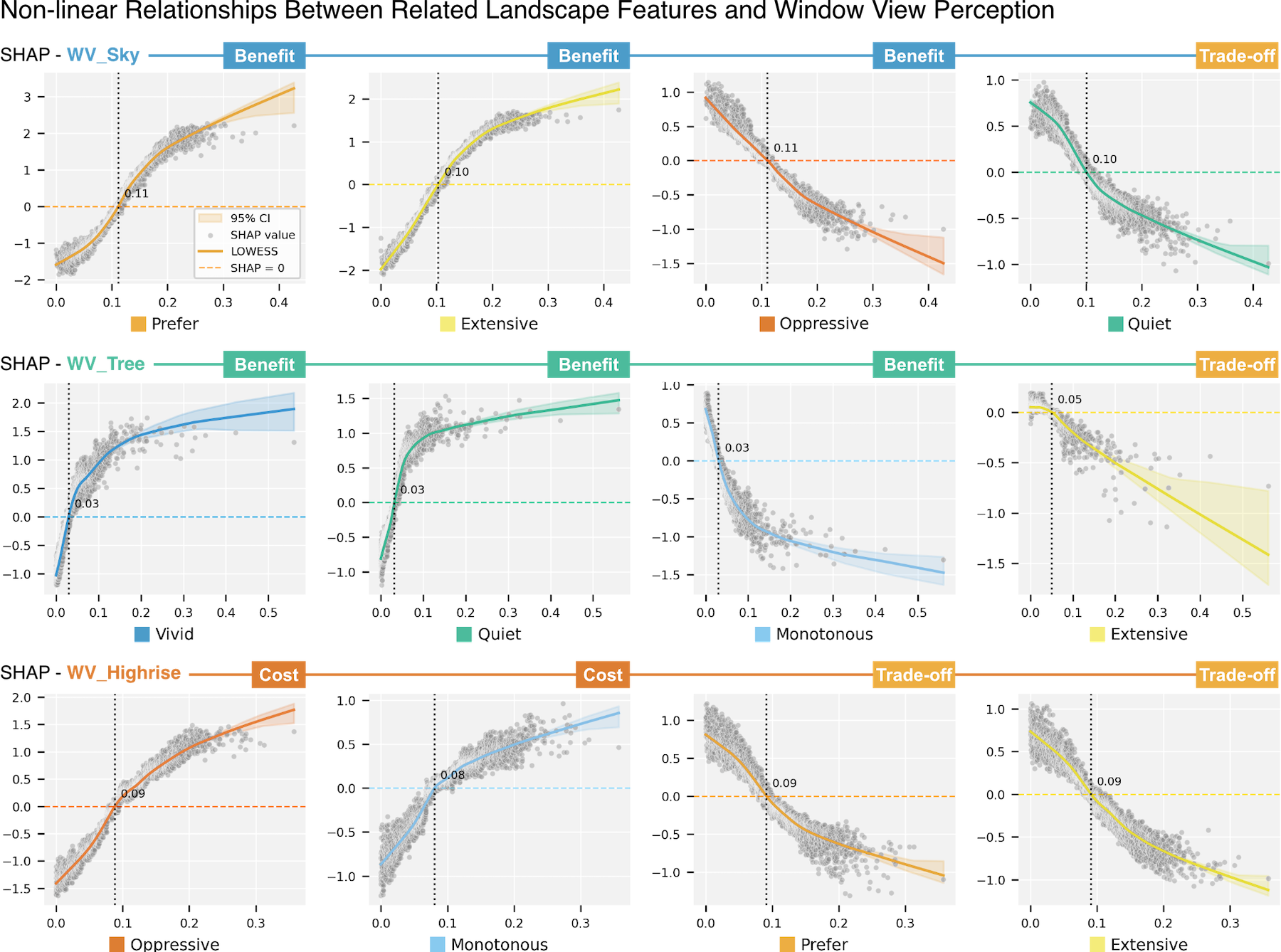}
    \caption{Non-linear relationships between key city landscape features and window view perceptions. Rows correspond to the three view elements with the largest SHAP effect ranges (\textit{WV\_Sky}, \textit{WV\_Tree}, and \textit{WV\_Highrise}); columns show, for each element, the four perception dimensions it affects most strongly, ordered from benefits (left) to trade-offs (right). Panel headers indicate whether the element's modelled contribution constitutes a benefit, a cost, or a trade-off for the corresponding perception. Vertical dotted lines mark where the LOWESS-smoothed SHAP curve crosses zero, i.e.\ where the modelled contribution of the element switches sign.}
    \label{non-linear}
\end{figure}

Figure \ref{non-linear} further unpacks these relationships, tracing how the modelled contributions of the three view elements with the largest SHAP effect ranges (\textit{WV\_Sky}, \textit{WV\_Tree}, and \textit{WV\_Highrise}) evolve as their visible proportions increase. The LOWESS-smoothed SHAP curves are distinctly non-linear: the benefits of sky for Prefer and Extensive accumulate steadily but trade off against Quiet once sky exceeds roughly 10\% of the view frame; the positive contributions of trees to Vivid and Quiet rise steeply at low coverage (around 3--5\%) before saturating, while their penalty on Extensive deepens with additional coverage; and high-rise buildings increase Oppressive and Monotonous perceptions across their range, with their contributions to Prefer and Extensive switching from positive to negative at approximately 9\% of the view. The vertical dotted lines mark these sign-switch thresholds, which we revisit as candidate composition-based design references in the Discussion.

Additional landscape elements show specific perceptual associations: higher proportions of visible roads (\textit{WV\_Road}) are associated with lower Quiet perception, possibly reflecting the co-occurrence of visible transportation infrastructure with denser, noisier urban environments. Grass (\textit{WV\_Grass}) is associated with higher Vivid and Quiet perceptions, suggesting that ground-level vegetation may accompany greater visual liveliness and environmental serenity. Among built environment variables, higher building density (\textit{BD}) and greater building height variation (\textit{BH\_SD}) are associated with Monotonous and Oppressive perceptions, consistent with dense, varied urban morphology co-occurring with more negative psychological responses. Finally, NDVI is associated with slightly lower oppressive feelings, while higher NDBI is associated with higher Monotonous perception, pointing to the contrasting correlates of vegetation density versus built environment intensity.

\section{Discussion}

\subsection{Implications for urban design and planning practice}

The findings offer empirical support for urban design and planning. We stress that these computational results are intended to inform rather than dictate design decisions: they offer a strong empirical support for design intuition, not a deterministic rulebook. The identification of optimal city landscape compositions and horizontal-vertical perception patterns offers actionable insights for creating psychologically supportive environments for residents, which should be weighed alongside the cultural, regulatory, programmatic, and aesthetic considerations that fall outside the scope of any urban computational model.

Given the demonstrated positive impacts of window views on psychological well-being, several established built environment evaluation standards have incorporated window view assessment as a criterion \citep{abdelrahman2023visible}. The current consolidated edition of the European daylight standard, EN 17037:2018+A1:2021 (\textit{Daylight in Buildings}), includes a dedicated assessment of view out based on the dimensions of the view opening, horizontal sight angle, outside viewing distance, the number of visible view layers —-- sky, landscape, and ground --— and the quality of the environmental information provided by the view \citep{en17037}. The current WELL Building Standard v2 addresses connections to nature under its Mind concept, particularly through features concerning access and enhanced access to nature, which recognise visual as well as physical contact with natural elements \citep{wellv2}. In LEED v5 for Building Design and Construction, \textit{Quality Views} is incorporated as a pathway within the EQc2 \textit{Occupant Experience} credit. It awards points where views of an outdoor natural or urban environment are provided to at least 75\% or 90\% of the regularly occupied floor area and specifies additional requirements concerning glazing, visible content, viewing distance, and occupant proximity to the view \citep{leedv5}. Similarly, BREEAM International New Construction Version 7 assesses \textit{view out} within its Hea,01 \textit{Natural Light} issue, alongside its daylight and glare-control provisions \citep{breeamv7}. Although these schemes specify various requirements concerning access to views, sightlines, viewing distance, glazing, visible layers, or broad categories of view content, they generally do not prescribe quantitative thresholds for the proportional composition of individual visual elements, such as the respective shares of buildings, vegetation, and sky within the window view. Equally important, the majority of these standards operate at the scale of the individual building or dwelling, specifying view-quality criteria assessed per window or per regularly occupied space; they are not designed to support assessment or benchmarking at the urban scale. Although their per-window criteria can in principle be extended to city-wide contexts, doing so requires bridging this scale gap --- precisely the gap that our citywide, imagery-based framework is intended to address. Our findings provide candidate reference points that could help inform such composition-based metrics. In our Wuhan dataset, the modelled contribution of visible high-rise buildings to perceived preference and openness switched from positive to negative once high-rise facades exceeded roughly one tenth of the panoramic view frame --- approximately one quarter of the average exterior view, as assessed from a viewpoint positioned one metre from the window --- and deteriorated steadily thereafter (Figure \ref{non-linear}). Similarly, the benefits of visible sky for preference and openness came at the cost of perceived quietness once sky exceeded a comparable share of the view. These values may be interpreted as indicative design references rather than fixed thresholds. 

Strategic integration of the city landscape into window views should maintain balanced compositions across window views at different floor levels. Urban designers and planners should consider sight-line methodology to reserve sight corridors that preserve visual access to the sky and distant features from buildings while incorporating diverse natural elements at multiple urban scales with methodologies such as vertical greening and pocket parks (the indicative design strategy map introduced below). The differential impacts of high-rise buildings versus low-rise buildings suggest opportunities for thoughtful urban design strategies. Planning urban sight corridors, strategically placing low-rise structures, and varying building heights can mitigate the visual monotony associated with high-rise building clusters \citep{wang2024assessing}. In dense urban contexts where high-rise buildings are necessary, vertical greening strategies --- including green walls, vegetated balconies, and integrated landscape features on buildings --- can enhance visual preference and reduce potential oppressive sensations \citep{chung2022study}.

Urban designers should adopt integrated approaches that combine natural and architectural elements. This method should focus on creating window views with diverse and visually appealing city landscapes while reducing monotonous and oppressive building arrangements, which may help improve residents' life satisfaction and mental well-being in dense urban areas. To demonstrate how these implications can be operationalised at the city scale, we translated the predicted perception scores and visible-environment composition into six complementary design-support maps, assembled in the upper panel of Figure \ref{fig_discussion} (computational details in the Appendix). The set is designed to answer distinct planning questions: a \textit{design-priority index} locates where negative window-view experience concentrates; an \textit{indicative design-strategy} map assigns each hexagon the most actionable intervention type (vertical or facade greening, additional vegetation and pocket parks, road buffering and screen planting, or preservation of sight corridors) according to its dominant visible-environment issue; a \textit{vertical-greening priority} map highlights areas where visible high-rise massing, oppressive sensations, and scarce visible greenery coincide; a \textit{view-quality inequality} map exposes neighbourhoods where a favourable average conceals large disparities between dwellings; a \textit{low-floor quality deficit} map identifies where lower floors lag substantially behind upper floors and may warrant low-level planting and sight-line design; and a composite \textit{view-quality index} provides a citywide benchmark distinguishing priority cold spots from reference exemplars. Consistent with our positioning of the framework as decision support, these maps indicate \textit{where} and \textit{what type} of intervention may be most beneficial, rather than prescribing specific design solutions; they are intended as a spatial evidence layer to be combined with the situated knowledge of designers, planners, and communities.

Taken together, the implications above are best understood as computational evidence that complements, rather than replaces, professional design judgement. Our framework can surface human-centred perceptual tendencies at an urban scale and flag where window-view quality may be compromised, but translating these signals into specific interventions and planning actions still requires the situated expertise of urban designers and planners, who must reconcile them with site context, regulatory constraints, and stakeholder values. We therefore position this work as a human-centred decision-support tool that strengthens, but does not supplant, the creative and contextual reasoning at the heart of the design process.

\begin{figure}[H]
    \centering
    \includegraphics[width=1\textwidth]{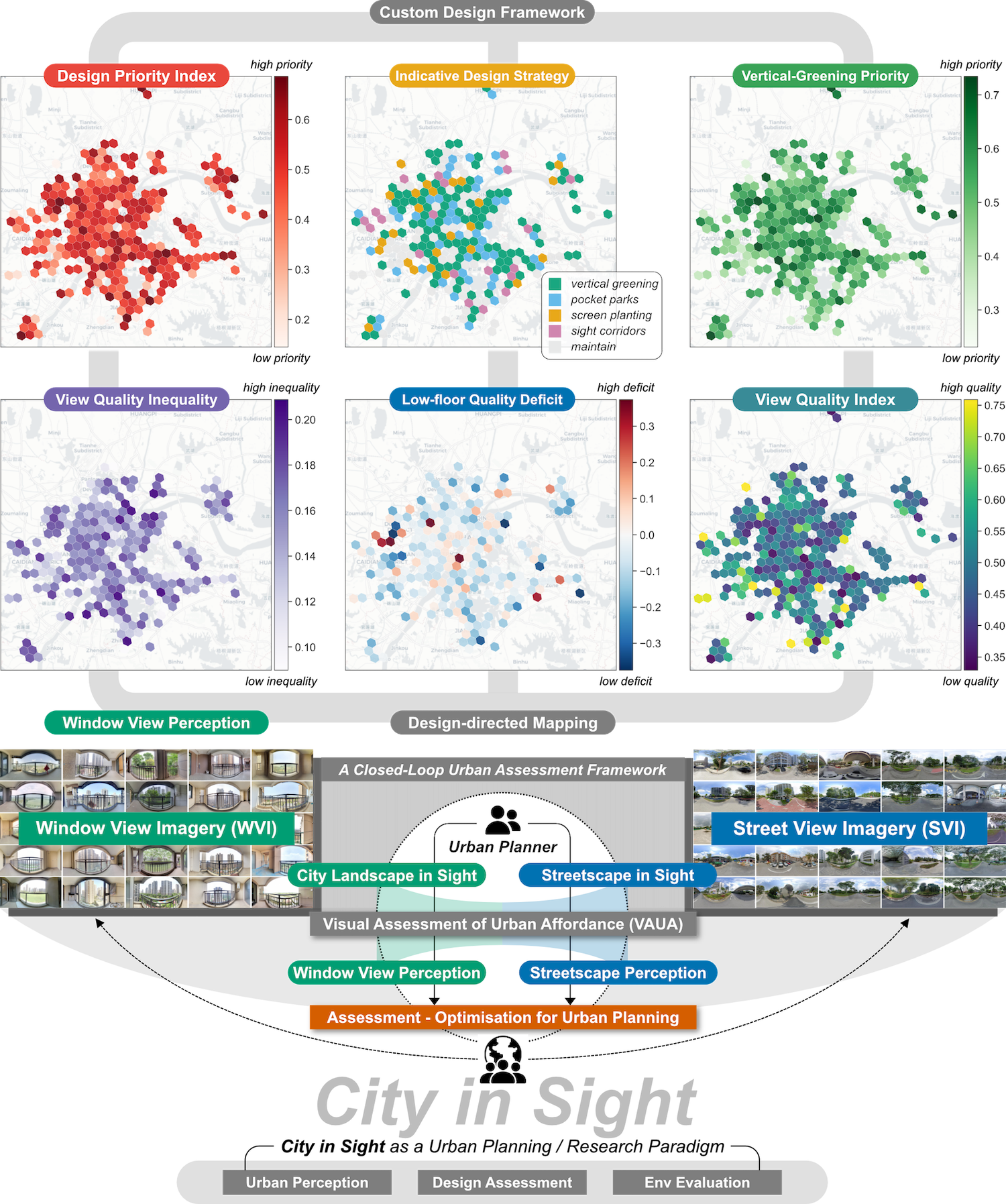}
    \caption{Window view perception driven urban design-support mapping (top) and the "City in Sight" conceptual framework integrating WVI and SVI for comprehensive urban view assessment for planning insights and applications (bottom).}
    \label{fig_discussion}
\end{figure}

Beyond the design-directed mapping in its upper panel, Figure \ref{fig_discussion} (bottom) illustrates a broader "City in Sight" paradigm that contextualises this research within a comprehensive visual assessment of urban affordance (VAUA) framework, where urban affordance refers to how the built environment supports human perception and social activities. This figure proposes that WVI complements SVI to create a holistic approach for visual assessment of the urban environment. While SVI captures streetscape perceptions from pedestrian perspectives in public spaces, WVI provides complementary insights into window views from interior viewpoints in residential buildings --- representing the private, intimate experience of city landscapes from within residents' homes. Together, these two imagery types form the foundation of the VAUA framework, enabling urban designers and planners to consider both public space experiences and private residential visual quality in integrated urban planning decisions, to better support social dynamics in future cities. This closed-loop framework supports evidence-based urban design and planning that addresses the full spectrum of human living experiences in the city.

\subsection{Limitations and future research directions}

Some limitations need further discussion and future work. First of all, data collection in this study was limited to Wuhan city, which may raise questions about the generalisability to other cities with different urban morphologies and cultural contexts. Future research should expand to more cities for diverse urban contexts, balancing comprehensiveness with the practical challenges of multi-city data acquisition~\citep{peng2023impact}.

Secondly, the real estate imagery underlying this study captures each view under fixed --- and systematically favourable --- atmospheric and temporal conditions. Because such imagery is produced for marketing purposes, it carries a selection bias: agents tend to photograph listings under favourable weather, optimal lighting, and advantageous camera angles, so images captured on rainy, hazy, or poorly lit days rarely appear in listings. This bias is likely to elevate the baseline scores of the positively valenced dimensions --- particularly Prefer and Extensive --- relative to residents' everyday lived experience. Two factors partially mitigate this effect: our dataset draws on standardised 360-degree panoramic balcony tours on Lianjia platform rather than free-form photography, which constrains photographer discretion; and the large, spatially diverse sample (12,334 images from 1,377 complexes across urban and suburban Wuhan) attenuates the influence of any individual listing decision on the aggregate, citywide findings. The fixed capture conditions also mean that, although the non-immersive VR platform provides more realistic window view experiences than traditional static image questionnaires, the perception survey cannot reflect the dynamic elements of reality --- moving clouds, vegetation changes across seasons, varied weather conditions, and daily sunlight transitions --- that are known to influence human perceptions of window views \citep{rodriguez2021subjective, moscoso2021window, ko2022window}. The absolute perception levels should therefore be read as upper-bound tendencies, and future studies should incorporate WVIs spanning multiple seasons, times of day, and weather conditions --- for example through video-based VR experiences or repeated captures over time --- to reduce this atmospheric and temporal selection bias \citep{wang2024assessing, sharam2023design}. Accordingly, our results should be interpreted as perceptions derived from the window views that are available through real estate listing imagery, rather than as a complete representation of all residential window views experienced by Wuhan's residents.

Thirdly, our perception labels were crowdsourced from 304 participants, a sample size that is reasonable for the pairwise comparison design. However, we did not collect detailed demographic information from participants, and perceptions of window views may vary with age, income, residential experience, cultural background, and housing preference, as recent large-scale work on street-view perception has shown that urban visual perception varies systematically across demographics and personalities \citep{quintana2025global}. The reported perceptions should therefore be read as an aggregate signal that may not capture systematic differences across population subgroups; collecting demographic covariates to model such perceptual heterogeneity --- for example, how preferences differ across age groups or housing situations --- is an important direction for future work.

Fourthly, the current framework focuses on six key perceptual dimensions, but additional aspects also need further investigation. More urban perception topics --- such as privacy, complexity, seasonal bias, and city landscape preferences across cultures --- could provide a more comprehensive understanding of window view experiences \citep{ogawa2024evaluating, zhao2025quantifying, quintana2025global}.

Fifthly, all WVIs in this study represent living room balcony views, a deliberate restriction that ensures functional consistency across samples and minimises privacy-related confounds. Window views experienced from other room types --- such as bedroom windows, where privacy considerations may temper residents' desire for visual connection to the outside --- are therefore not represented in the dataset. Since perceptual responses may vary with room function and its associated expectations, future studies should examine how window view perceptions differ across room types, for example between communal living spaces and more private bedroom settings.

Sixthly, this study focuses exclusively on view content and does not account for view access (e.g., window size, orientation, sill height, and shading conditions) or view clarity (e.g., glazing quality, condensation, and physical obstructions). Even highly desirable view content may be diminished in practice by poor view access or clarity. Future work should integrate these complementary dimensions for a more holistic assessment of window view quality with real estate imagery \citep{ko2021}.

Finally, the inference models and their SHAP attributions should be interpreted as associations rather than causal effects. SHAP quantifies how each predictor contributes to a model's output given the observed data, but the analysis is cross-sectional and observational, and the built-environment predictors are mutually correlated (e.g.\ the compositional window-view and land-cover proportions). Consequently, the directions and magnitudes we report reflect statistical relationships learned by the model, not the outcome of controlled interventions; an apparent effect of one variable may partly stand in for co-varying environmental or socio-spatial factors that we did not measure. We therefore interpret individual SHAP attributions with caution, mindful of the residual collinearity among predictors (Appendix, Figure \ref{fig_vif}), and frame the design implications as hypotheses to be confirmed by longitudinal or experimental studies that manipulate window-view composition directly.

Moreover, future research could benefit from integrating physiological measurement techniques and equipment (eye-tracking devices, EEG, stress indicators) with perceptual surveys to validate human psychological responses ~\citep{liu2025mapping, yang2026spatial} and incorporate more advanced computational methods, such as semantic urban scene understanding ~\citep{ito2025zensvi}, 3D or network-based city modelling analysis ~\citep{yap2023urbanity, fan2025imagebased, fujiwara2026voxcity}, and multimodal sensory integration ~\citep{fujiwara2024microclimate, cheng2025walking, yang2025thermal}. The development of real-time assessment tools and integration with urban digital twin platforms also represents promising directions for future practical applications ~\citep{lei2025developing, lei2026multidimensional}. Furthermore, integrating LLM-based reasoning-capable AI agents \citep{yang2025reasoning} could enable more sophisticated decision-making frameworks that transparently evaluate trade-offs among visual quality of the city landscape, regulatory constraints, and stakeholder preferences in urban design and planning applications.

\section{Conclusions}

This study establishes a comprehensive, scalable, and fully crowdsourced framework for modelling and understanding human perceptions of urban window views. By integrating crowdsourced real estate imagery  ---  uploaded by property agents and sellers  ---  with crowdsourced perceptual assessments and a hybrid neural network for perception modelling, we demonstrate the potential of a dual-crowdsourced framework (multiple urban data streams) for window view perception research. By analysing 12,334 real WVIs from Wuhan city, China, combined with 27,477 valid pairwise visual-perceptual comparisons across six subjective perception dimensions, we developed and validated hybrid neural network models capable of predicting multidimensional perception scores at an urban scale.

The research reveals systematic horizontal-vertical patterns in window view perception, reflecting impacts of urban environmental characteristics. Spatial autocorrelation analysis reveals significant geospatial clustering in window view perception, with vivid showing the strongest dependence. Floor level influences perceptions of window views: higher floors are associated with greater openness and preference, while lower floors offer greater quiet and vividness.

Critically, the study identifies non-linear relationships between city landscape elements and human psychological responses, challenging assumptions about the monotonic benefits of natural features on human perception. Sky visibility and vegetation exhibit clear trade-offs --- expanded sky views diminish perceived quietness, and dense tree cover curtails perceived openness by blocking distant sight lines --- while the contribution of visible high-rise buildings to preference and openness turns negative once they exceed roughly one tenth of the view frame (about one quarter of the exterior view), underscoring the importance of balanced city landscape compositions in urban views rather than maximising individual components.

The SHAP-based interpretation framework provides actionable insights for urban design practice, suggesting indicative quantitative reference points that could inform built environment evaluation standards, strategies for vertical greening in dense developments, and approaches for creating diverse yet harmonious city landscapes. The identification of three distinct window view perception types --- preferred-expansive views (Type 1), monotonous-oppressive views (Type 2), and quiet-vivid views (Type 3) --- provides a typological framework for understanding and designing residential visual environments.

This framework demonstrates the feasibility and value of integrating fully crowdsourced data --- both WVIs and their perceptual labels from online surveys --- with deep learning methodologies for comprehensive subjective window view experience assessment at the urban scale. The approach provides a replicable method for future window view perception analysis, with potential applications in data-driven urban design, urban planning, window view quality evaluation, and urban digital twins. Furthermore, it gives more attention to geo-tagged images obtained from real estate ads, an emerging user-generated data source for urban analytics. Future research should expand the geographic and cultural scope, incorporate temporal dynamics, and integrate additional perceptual dimensions to enhance the framework's comprehensiveness and applicability across diverse urban environments. 

\section*{CRediT authorship contribution statement}

Chucai Peng: Conceptualisation, Methodology, Formal analysis, Software, Writing - original draft. Sijie Yang: Conceptualisation, Methodology, Formal analysis, Investigation, Software, Visualisation, Data curation, Writing - original draft. Ang Liu: Investigation, Data curation. Yang Xiang: Investigation, Software. Zhixiang Zhou: Validation, Funding acquisition, Supervision. Filip Biljecki: Conceptualisation, Writing - review \& editing, Supervision.

\section*{Acknowledgements}

We gratefully acknowledge the participants of the experiment. We thank our colleagues at the NUS Urban Analytics Lab for the discussions. This research is supported by the China Scholarship Council (grant No. 202306760114) and NUS Research Scholarship (NUSGS-CDE DO IS AY22\&L GRSUR0600042). This research is part of the project, Large-scale 3D Geospatial Data for Urban Analytics, which is supported by the National University of Singapore under the Start-Up Grant R-295-000-171-133.

\section*{Data availability}

To support reproducibility, we openly release the analysis pipeline, the derived data, the processed imagery, and the trained models across two repositories. The code repository (\url{https://github.com/Sijie-Yang/City-Landscape-In-Sight}) provides: (i) the complete feature-extraction and modelling code as documented notebooks (perception survey processing, image pre-processing and visual-feature extraction, spatial sampling, perception prediction modelling, data analytics, inference modelling, and design-support mapping), together with the image pre-processing scripts; (ii) the image-derived feature tables for both the surveyed and the citywide WVIs; (iii) the nested H3 spatial units (hexagon and complex identifiers) and the cross-validation fold assignments; and (iv) the anonymised perception data, comprising the pairwise comparison responses, the TrueSkill perception scores, and the citywide predicted perception scores. The accompanying dataset repository on Hugging Face (\url{https://huggingface.co/datasets/sijiey/City-Landscape-In-Sight}) hosts the larger binary artefacts: the processed window view images (the 499 surveyed and the 12,334 citywide WVIs) and the trained model weights for all six perceptual dimensions. The raw window view images obtained from the listing platform are not redistributed owing to platform licensing restrictions; however, the processing scripts required to regenerate the processed imagery are provided.

\section*{Declaration of generative AI and AI-assisted technologies in the writing process}

We utilised Grammarly and ChatGPT for language refinement and grammar checks throughout the writing process. After using this tool/service, the authors reviewed and edited the content as needed and took full responsibility for the publication's content. While these tools helped ensure language accuracy and clarity, the authors generated all scientific insights, conclusions, and content. The final manuscript underwent thorough review and editing by the authors to ensure accuracy and integrity.

\appendix
\renewcommand{\thefigure}{A.\arabic{figure}} 
\setcounter{figure}{0} 
\renewcommand{\thetable}{A.\arabic{table}}
\setcounter{table}{0}
\renewcommand{\theequation}{A.\arabic{equation}}
\setcounter{equation}{0}

\section*{Appendix}
\label{sec:sample:appendix}

\subsection*{Spatial statistics}

The Global Moran's I statistic used to quantify the overall spatial autocorrelation of each perception dimension is defined as
\begin{equation}
    \label{eq:moran}
    I = \frac{n}{W}\,\frac{\sum_{i}\sum_{j} w_{ij}\,(x_i-\bar{x})(x_j-\bar{x})}{\sum_{i}(x_i-\bar{x})^2},
\end{equation}
where $n$ is the number of locations, $x_i$ is the mean perception score at location $i$, $\bar{x}$ is the global mean, $w_{ij}$ is the spatial weight between locations $i$ and $j$, and $W=\sum_{i}\sum_{j} w_{ij}$. The expectation under spatial randomness is $\mathbb{E}[I]=-1/(n-1)$.

The local Getis-Ord $G_i^*$ statistic used to identify hot spots and cold spots is defined as
\begin{equation}
    \label{eq:gistar}
    G_i^* = \frac{\sum_{j} w_{ij}\,x_j - \bar{x}\sum_{j} w_{ij}}{S\,\sqrt{\dfrac{n\sum_{j} w_{ij}^{2}-\left(\sum_{j} w_{ij}\right)^{2}}{n-1}}},
    \qquad S=\sqrt{\frac{\sum_{j}(x_j-\bar{x})^2}{n}},
\end{equation}
where the weights $w_{ij}$ are derived from the six nearest hexagonal neighbours of cell $i$, including the focal cell itself ($w_{ii}=1$). The resulting $G_i^*$ is a z-score whose sign and magnitude indicate significant high-value (hot spot) or low-value (cold spot) clusters.

\subsection*{Detailed model performance}

Table \ref{tab:appendix_best_metrics} reports the complete per-split performance of the best hybrid NN checkpoint for each perception dimension, providing the detailed $R^2$, MSE, and RMSE values summarised in Figure \ref{fig_model_prediction_best}.

\begin{table}[htb]
    \centering
    \caption{Detailed performance of the best hybrid NN checkpoint (selected on validation $R^2$) for the six window view perception dimensions across the training, validation, and test splits.}
    \label{tab:appendix_best_metrics}
    \resizebox{\textwidth}{!}{%
    \begin{tabular}{lccccccccc}
        \toprule
        & \multicolumn{3}{c}{$R^2$} & \multicolumn{3}{c}{MSE} & \multicolumn{3}{c}{RMSE} \\
        \cmidrule(lr){2-4} \cmidrule(lr){5-7} \cmidrule(lr){8-10}
        Dimension & Train & Val & Test & Train & Val & Test & Train & Val & Test \\
        \midrule
        Prefer     & 0.885 & 0.695 & 0.773 & 0.2435 & 0.5680 & 0.5093 & 0.4934 & 0.7537 & 0.7137 \\
        Monotonous & 0.842 & 0.569 & 0.560 & 0.3526 & 0.8936 & 0.7041 & 0.5938 & 0.9453 & 0.8391 \\
        Quiet      & 0.813 & 0.504 & 0.574 & 0.3766 & 1.1658 & 0.8567 & 0.6137 & 1.0797 & 0.9256 \\
        Extensive  & 0.874 & 0.760 & 0.743 & 0.2612 & 0.5307 & 0.5185 & 0.5111 & 0.7285 & 0.7201 \\
        Vivid      & 0.867 & 0.636 & 0.668 & 0.2749 & 0.6725 & 0.7704 & 0.5243 & 0.8201 & 0.8777 \\
        Oppressive & 0.872 & 0.710 & 0.739 & 0.2494 & 0.6497 & 0.5944 & 0.4994 & 0.8060 & 0.7709 \\
        \bottomrule
    \end{tabular}%
    }
\end{table}

\subsection*{Spatial cross-validation comparison}

Table \ref{tab:appendix_kfold_comparison} reports the complete per-dimension comparison of the full hybrid NN under random stratified and spatial block K-fold cross-validation, corresponding to Figure \ref{fig_kfold_comparison}.

\begin{table}[htb]
    \centering
    \caption{Mean fold test $R^2$ of the full hybrid NN under random stratified versus spatial block K-fold cross-validation for the six window view perception dimensions.}
    \label{tab:appendix_kfold_comparison}
    \begin{tabular}{lccc}
        \toprule
        Dimension & Random stratified & Spatial block & $\Delta$ (random $-$ spatial) \\
        \midrule
        Prefer     & 0.7578 & 0.7482 & $+0.0096$ \\
        Monotonous & 0.4851 & 0.4906 & $-0.0055$ \\
        Quiet      & 0.5047 & 0.5048 & $-0.0001$ \\
        Extensive  & 0.7072 & 0.6897 & $+0.0175$ \\
        Vivid      & 0.6281 & 0.5620 & $+0.0661$ \\
        Oppressive & 0.6494 & 0.6434 & $+0.0060$ \\
        \bottomrule
    \end{tabular}
\end{table}

\subsection*{Multicollinearity diagnostics}

\begin{figure}[H]
    \centering
    \includegraphics[width=1\textwidth]{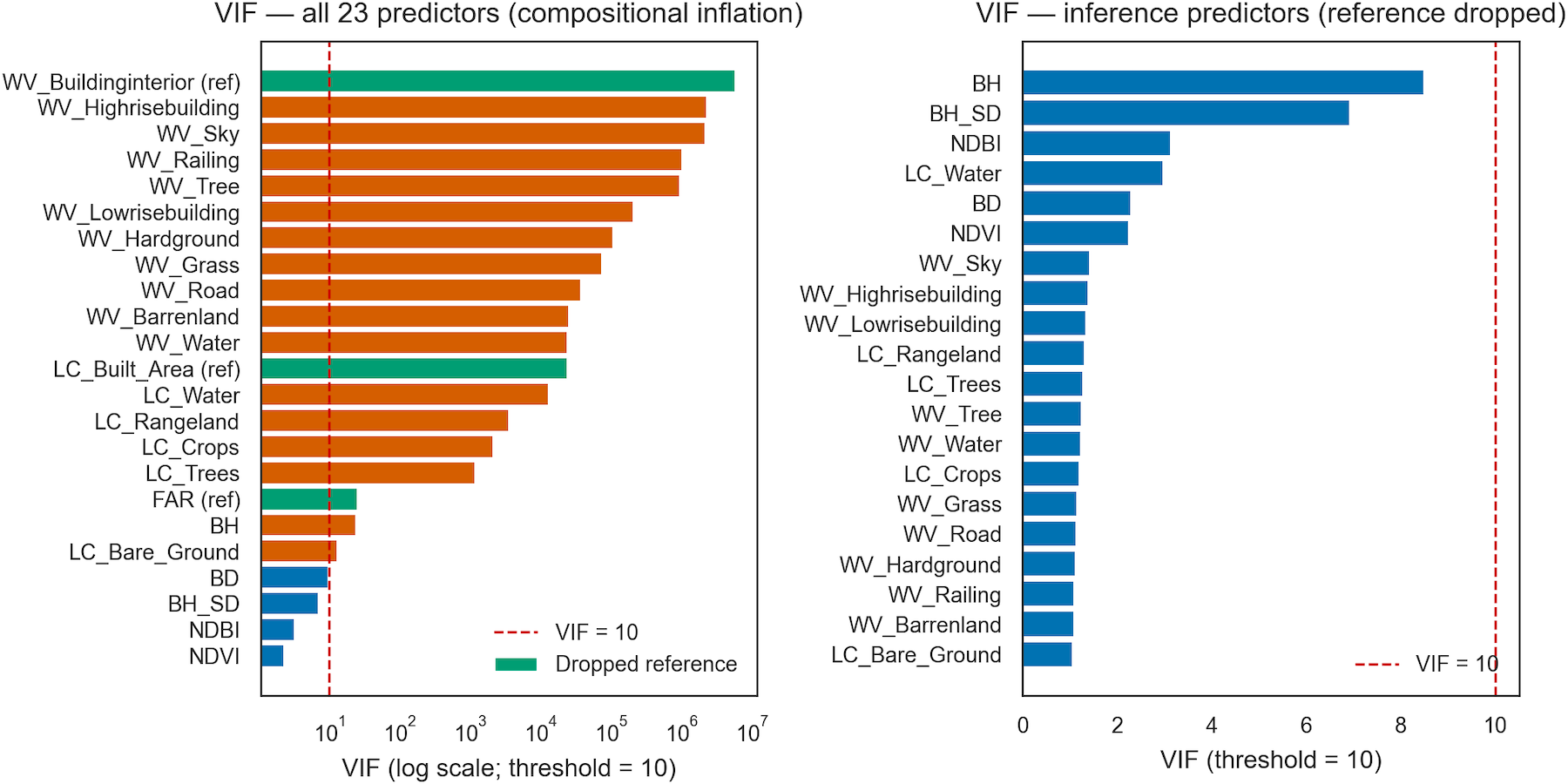}
    \caption{Variance inflation factors (VIF) for the inference predictors. Left: all 23 predictors (log scale), where the compositional window-view and land-cover proportions inflate the VIF; the three dropped reference variables (\textit{WV\_Buildinginterior}, \textit{LC\_Built\_Area}, \textit{FAR}) are highlighted. Right: the reference-dropped predictor set, for which all VIF fall below the threshold of 10.}
    \label{fig_vif}
\end{figure}

Figure \ref{fig_vif} reports the VIF values used to diagnose multicollinearity among the inference predictors. Because the window-view and land-cover classes are compositional (each group's proportions sum to one), the raw VIF of the full 23-predictor set is structurally inflated to extreme values (left panel, log scale). Following standard practice for compositional data, we dropped one reference category from each group---the dominant class for the window-view (\textit{WV\_Buildinginterior}) and land-cover (\textit{LC\_Built\_Area}) proportions, and the most collinear urban-form variable (\textit{FAR})---before recomputing the VIF with an intercept. After this correction, all retained predictors have VIF below the conventional threshold of 10 (right panel; maximum $\approx 8.5$ for building height), indicating that the residual collinearity is mild. As a further safeguard, the inference models we employ (regularised linear models, PLS, and tree-based methods) are robust to multicollinearity, and we interpret individual SHAP attributions with this residual collinearity in mind, rather than as the fully isolated effects of single predictors.

\subsection*{SHAP attribution scatter plots}

\begin{figure}[H]
    \centering
    \includegraphics[width=1\textwidth]{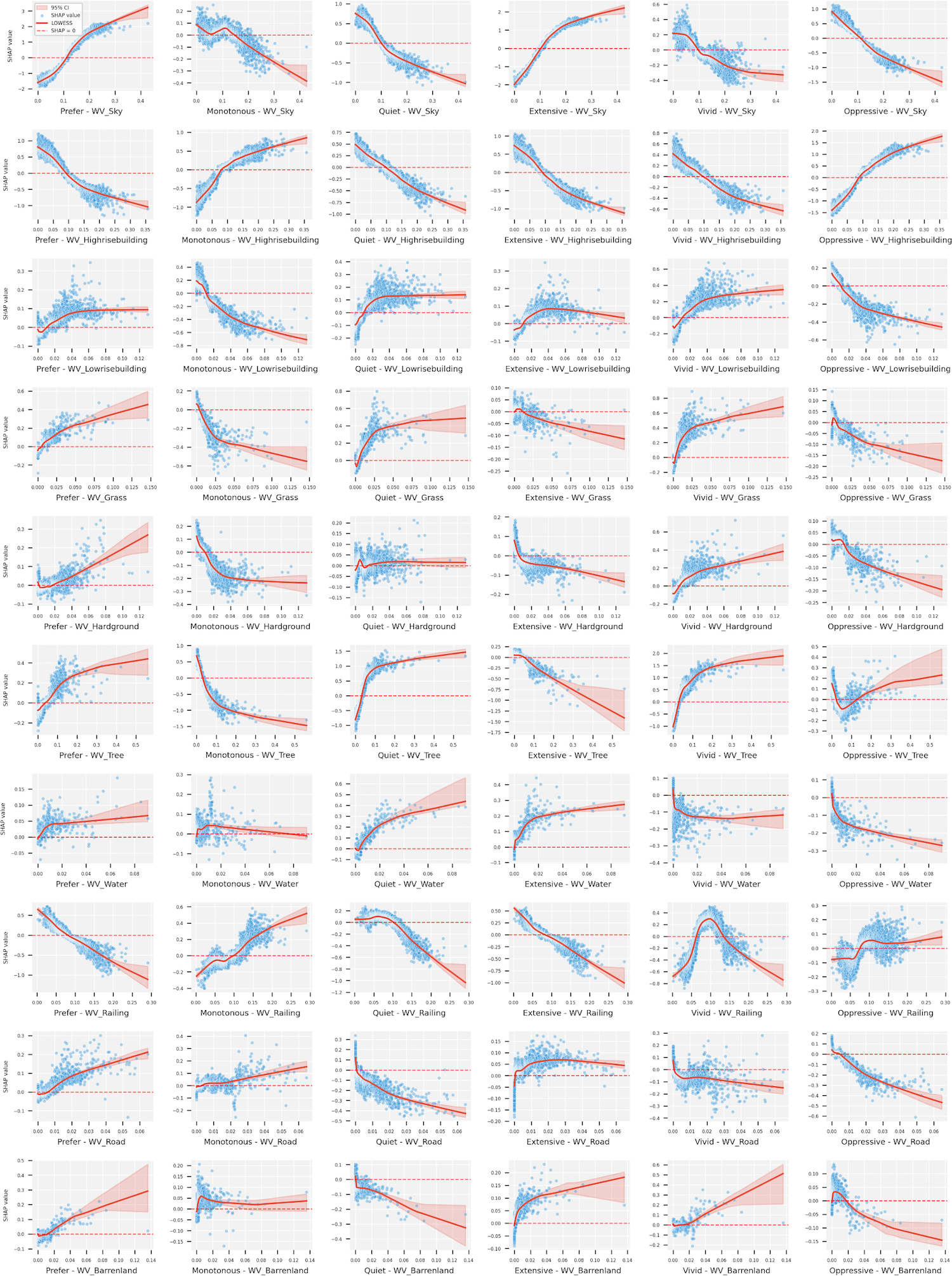}
    \caption{SHAP scatter plots of window view variables and perception.}
    \label{appendix_fig_wv}
\end{figure}

\begin{figure}[H]
    \centering
    \includegraphics[width=1\textwidth]{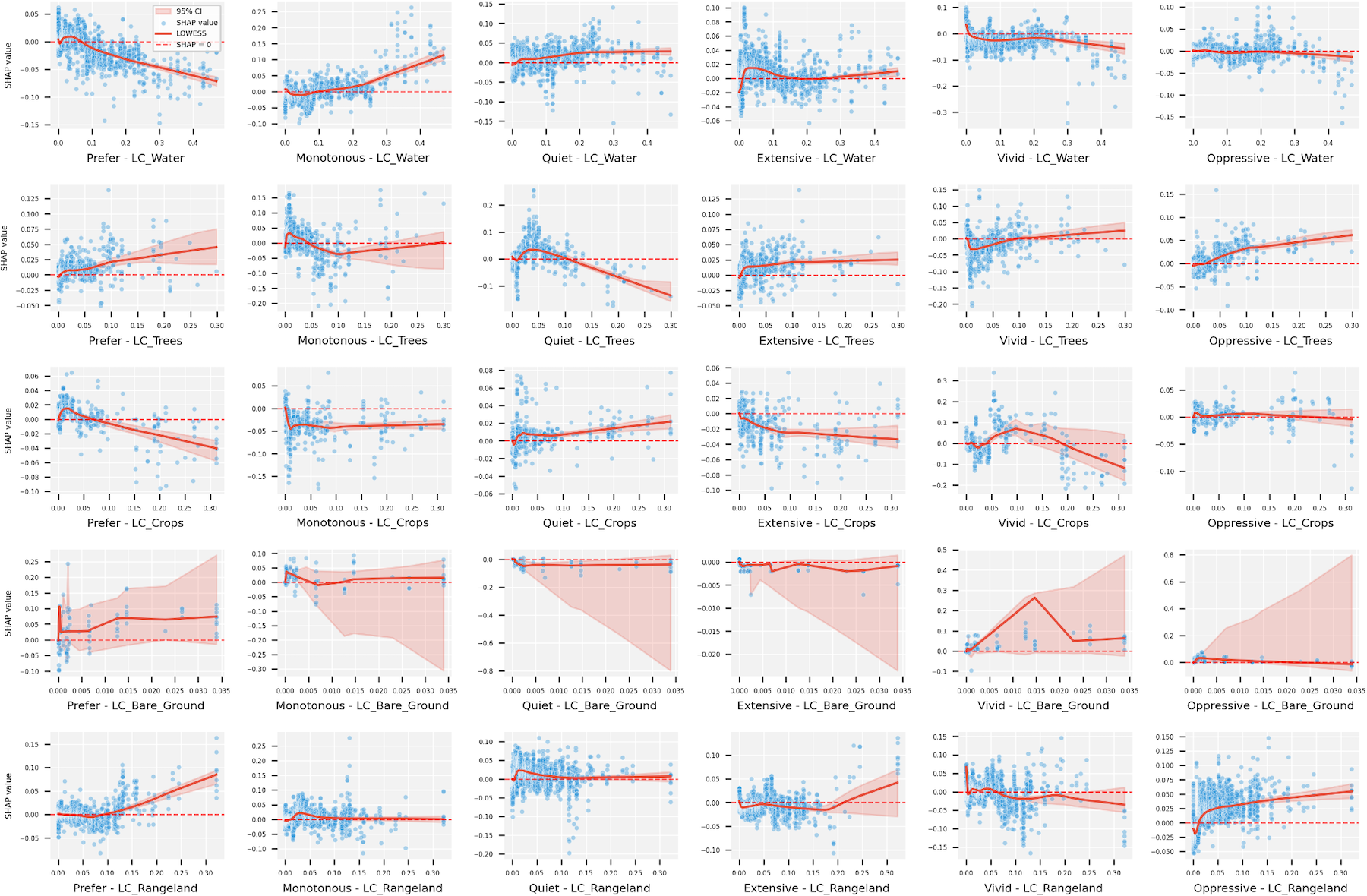}
    \caption{SHAP scatter plots of land cover variables and perception.}
    \label{appendix_fig_lc}
\end{figure}

\begin{figure}[H]
    \centering
    \includegraphics[width=1\textwidth]{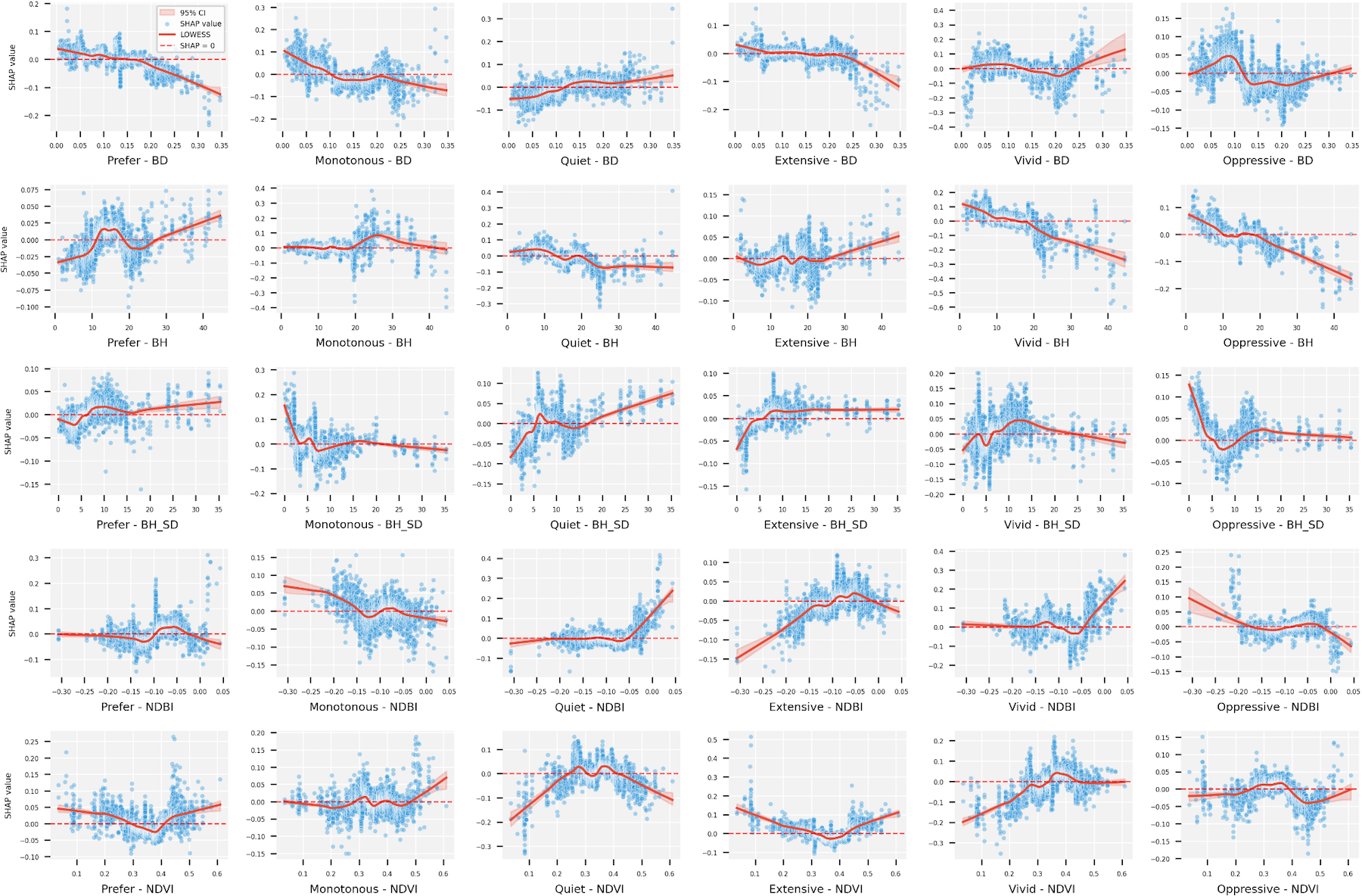}
    \caption{SHAP scatter plots of building form variables and perception.}
    \label{appendix_fig_bf}
\end{figure}

Figures \ref{appendix_fig_wv}--\ref{appendix_fig_bf} present detailed SHAP scatter plots with LOWESS-smoothed trends, illustrating the relationships between individual predictors and perception scores across all six dimensions, grouped by window view, land cover, and building form variables.

\subsection*{Design-support mapping algorithms}

This section details the computation of the six design-support maps shown in the upper panel of Figure \ref{fig_discussion}. All maps are built on the H3 hexagonal grid (resolution 7): each WVI is assigned to the hexagon containing its complex coordinates, and the predicted perception scores together with the visible-environment composition are averaged per hexagon. Unless stated otherwise, $\tilde{x}$ denotes the min--max normalisation of a hexagon-level variable $x$ across all hexagons, $\tilde{x}=(x-x_{\min})/(x_{\max}-x_{\min})$, and colour scales are clipped to the 2nd--98th percentile of the mapped values.

\textit{(a) Design-priority index.} The concentration of negative window-view experience, computed per hexagon as the average of the normalised Oppressive and Monotonous scores and the inverted normalised Prefer score:
\begin{equation}
    \label{eq:design_priority}
    P \;=\; \tfrac{1}{3}\left(\widetilde{\mathit{Opp}} + \widetilde{\mathit{Mon}} + \bigl(1-\widetilde{\mathit{Pre}}\bigr)\right),
\end{equation}
so that higher values flag areas warranting greater design attention.

\textit{(b) Indicative design strategy.} A rule-based classification assigns each hexagon the first matching strategy in the following sequence, where all comparisons are against the citywide median (denoted $m(\cdot)$) of the hexagon-level values: (i) vertical/facade greening if visible high-rise exceeds $m(\mathit{WV\_Highrise})$ and Oppressive exceeds its median; (ii) add vegetation/pocket parks if visible trees fall below $m(\mathit{WV\_Tree})$ and either Monotonous is above or Vivid is below its median; (iii) road buffering/screen planting if visible road exceeds its median and Quiet is below its median; (iv) preserve sight corridors/openness if visible sky and Extensive are both below their medians; otherwise (v) maintain (low priority). The rules are evaluated in this order, so each hexagon receives a single indicative strategy.

\textit{(c) Vertical-greening priority.} Analogous to Eq.~\eqref{eq:design_priority}, combining the normalised visible high-rise proportion, the normalised Oppressive score, and the inverted normalised visible-tree proportion, $G=\tfrac{1}{3}(\widetilde{\mathit{WV\_Highrise}}+\widetilde{\mathit{Opp}}+(1-\widetilde{\mathit{WV\_Tree}}))$, so that high values identify areas where green walls, vegetated balconies, and facade greening are most indicated.

\textit{(d) View-quality inequality.} To expose within-area disparities hidden by hexagon means, the composite view-quality index (VQI; Eq.~\eqref{eq:vqi} in item~\textit{f} below) is first computed for each individual WVI (scaling each dimension over its point-level range), and the inequality of a hexagon is the standard deviation of these point-level values across all WVIs it contains. Hexagons with fewer than five WVIs are excluded to ensure a stable dispersion estimate.

\textit{(e) Low-floor quality deficit.} The hexagon-level VQI (Eq.~\eqref{eq:vqi}) is computed separately for the low-floor and high-floor subsets of WVIs (floor categories as defined in the Method section), and the deficit is their difference, $\Delta=\mathrm{VQI}_{\mathrm{high}}-\mathrm{VQI}_{\mathrm{low}}$, evaluated only for hexagons sampled at both levels. Positive values mark areas where lower floors lag behind upper floors and may benefit most from low-level greening and sight-line design, whereas negative values indicate the reverse.

\textit{(f) Composite view-quality index.} A single index summarising all six perception dimensions. With each dimension scaled to $[0,1]$ over its observed range, the index averages the positively valenced dimensions and the complements of the negatively valenced ones:
\begin{equation}
    \label{eq:vqi}
    \mathrm{VQI} \;=\; \frac{1}{2}\left(\frac{1}{4}\sum_{c\,\in\,\{\mathit{Pre},\mathit{Qui},\mathit{Ext},\mathit{Viv}\}} \tilde{c} \;+\; \frac{1}{2}\sum_{c\,\in\,\{\mathit{Mon},\mathit{Opp}\}} \bigl(1-\tilde{c}\bigr)\right).
\end{equation}
Low values (cold spots) mark citywide priority zones, and high values reference exemplars of good window-view quality.

\bibliographystyle{elsarticle-harv} 
\bibliography{refs}





\end{document}